\pdfoutput=1

\documentclass[11pt]{article}

\usepackage[preprint]{acl}

\usepackage{times}
\usepackage{latexsym}
\usepackage{amsmath}

\usepackage{booktabs}
\usepackage{multirow}
\usepackage{tabularx}
\usepackage{graphicx} 
\usepackage{float} 

\usepackage[T1]{fontenc}

\usepackage[utf8]{inputenc}

\usepackage{microtype}

\usepackage{inconsolata}

\usepackage{graphicx}
\usepackage{array}
\usepackage{subcaption} 

\usepackage{booktabs}
\usepackage{multirow}
\usepackage{tcolorbox}

\usepackage{tcolorbox}

\usepackage{makecell}

\title{Can Competition Enhance the Proficiency of Agents Powered by Large Language Models in the Realm of News-driven Time Series Forecasting?}

\author{
 \textbf{Yuxuan Zhang\textsuperscript{1}},
 \textbf{Yangyang Feng\textsuperscript{1}},
 \textbf{Daifeng Li\textsuperscript{1}},
 \textbf{Kexin Zhang\textsuperscript{1}},
\\
 \textbf{Junlan Chen\textsuperscript{1}},
 \textbf{Bowen Deng\textsuperscript{1}},
\\
\\
 \textsuperscript{1}Sun Yat-sen University, Guangzhou, China
\\
 \small{
   \textbf{Correspondence:} \href{mailto:zhangyx528@mail2.sysu.edu.cn}{zhangyx528@mail2.sysu.edu.cn}
 }
}

\begin{document}
\maketitle
\begin{abstract}
Multi-agents-based news-driven time series forecasting is considered as a potential paradigm shift in the era of large language models (LLMs). The challenge of this task lies in measuring the influences of different news events towards the fluctuations of time series. This requires agents to possess stronger abilities of innovative thinking and the identifying misleading logic. However, the existing multi-agent discussion framework has limited enhancement on time series prediction in terms of optimizing these two capabilities. Inspired by the role of competition in fostering innovation, this study embeds a competition mechanism within the multi-agent discussion to enhance agents’ capability of generating innovative thoughts. Furthermore, to bolster the model’s proficiency in identifying misleading information, we incorporate a fine-tuned small-scale LLM model within the reflective stage, offering auxiliary decision-making support. Experimental results confirm that the competition can boost agents' capacity for innovative thinking, which can significantly improve the performances of time series prediction. Similar to the findings of social science, the intensity of competition within this framework can influence the performances of agents, providing a new perspective for studying LLMs-based multi-agent systems. The implementation code is available at \href{https://anonymous.4open.science/r/IA_news_model-D7D6/}{https://anonymous.4open.science/r/IA\_news\_model-D7D6/}.      
\end{abstract}

\section{Introduction}

Time series forecasting is a pivotal foundation for decision-making across a broad ranges of applications in economic, infrastructural, social domains \citep{liu2021pyraformer,xue2023promptcast, cao2023tempo}. The intent behind analyzing time series data is to detect the intricate and evolving inter-dependencies that characterize complex, dynamic real-world systems. Existing methods did not systematically connect complex social events with fluctuations in time series. Their ability to predict fluctuations in time series, such as sudden changes, is limited \citep{rasul2023lagllama, tang2024timeseriesforecastingllms}. 

News articles can provide crucial insights into unexpected incidents, policy changes, technological developments, and public sentiment shifts, which numerical data alone may not capture \citep{Rodrigues2019combinetsandtext,rasul2023lagllama,Wang2024,zhou2024one,Cheng2024sociodojo}. One direction for connecting news with time series is to transform the forecasting task into the prediction of the next token \citep{jin2023timellm,Wang2024}. This can better use the reasoning capabilities of LLMs \citep{gruver2024large}. However, the factors involved in this task encompass a wide range of knowledge, with complex correlations. An expansive landscape for strategic exploration, coupled with inherent uncertainties may amplify the reasoning errors \citep{Huang2025@survey}. For example, selecting the wrong news, or miscalculating the impact of the news will result in significant bias in the prediction results. Therefore, the key to improving this task lies in enabling the model to form a unique and effective mode of understanding the inner correlations between events and time series.

Multi-agent discussions can facilitate the formation of the desired mode by fostering diverse thinking and constructing better logics by reflections \citep{Liang2024@encourage,Wang2024@rethinking,Zhang2024@explore, Guan@2025mmdere}. However, these frameworks still have the \textbf{Degeneration-of-Thought (DoT) problem} \citep{Wang2024@rethinking,Liang2024@encourage}, which is the lack of novel thoughts due to the high confidence of the model after several rounds of discussions. Experimental findings indicate that these discussion frameworks do not yield significant enhancements when contrasted with single agents equipped with robust prompts \citep{Wang2024@rethinking}. In addition, the \textbf{Wrong Logic Propagation Error} \citep{Balepur2024@llmer,Wang2024@rethinking} that arises during discussions can also have a negative impact, because agents can be misled by information that appears to be correct due to the lack of auxiliary judgment methods.     

Drawing inspiration from the role of competition in fostering innovation within the stock market \citep{WANG201740}, we propose a hypothesis that can competition effectively address these two limitations? In the stock market, investors are able to perceive the loss brought by competition \citep{Chen2025@compet}. The competition awareness motivates investors to break the confidence in their original strategies and continuously innovate strategies to gain higher returns. Information asymmetry is a major factor leading to competition \citep{WANG201740}, investors will conceal their core strengths and speculate on the strategies of other competitors. Under that situation, agents can enhance their abilities to analyze and judge misleading information \citep{tampubolon2021informationasymmetrycompetitivemultiagent}, or spontaneously seek potential collaborators to gain a competitive advantage \citep{wu2024shallteamupexploring}. 

The competitive patterns of multi-agents in LLM interactions have already been studied, and most of the research focuses on social simulations in specific contexts, such as market competition \citep{zhao2024competeaiunderstandingcompetitiondynamics,wu2024shallteamupexploring} and game scenario modeling \citep{junprung2023exploringintersectionlargelanguage, wu2024shallteamupexploring, lan2024llmbasedagentsocietyinvestigation}. To date, research on improving task performance through competitive multi-agent systems remains scarce, which is crucial for leveraging LLMs to address core issues across various fields. The main contributions are summarized as follows.

\begin{itemize}
\item A competition mechanism is proposed to enhance agents' abilities in news-driven time series forecasting. Drawing on theories of competition and innovation, we incorporate Information Asymmetry, Competitive Awareness, and Survival of the Fittest into the framework of multi-agent collaborative discussion to investigate whether the competition can enhance the innovative thinking of agents, providing a new perspective for the optimization of LLM-based multi-agent systems.  

\item A multi-stage reflection (MSR) is designed to improve each agent’s analytical and judgment abilities by integrating a fine-tuned small LLM. MSR is important in stabilizing the operation of the competition mechanism and mitigating the wrong logic propagation error.

\item Experimental results show that the competition mechanism outperforms the baseline models, and the analysis of each agent's logic reveals that competitions can enhance the innovative thinking of agents. In addition, we observe the U-shape correlations between competitive intensity and agent's performances, in alignment with findings within the social sciences. This reflects LLM’s potential in simulating complex social activities.
\end{itemize}

\section{Related Work}

\subsection{LLMs for Time Series Forecasting}

LLMs have been widely applied to research in time series prediction tasks\citep{jin2023timellm, cao2023tempo, gruver2024large, rasul2023lagllama, zhou2024one}. Current research on enhancing LLMs for time series forecasting has focused on three primary approaches: model reprogramming\citep{xue2023promptcast, jin2023timellm, sun2023test, gruver2024large}, model fine-tuning\citep{cao2023tempo, das2023decoder, garza2023timegpt, rasul2023lagllama}, and incorporating contextual information\cite{tang2024timeseriesforecastingllms,jin2023timellm}. 

\citet{Wang2024} proposed a framework utilizing reasoning agents to filter relevant news and assist LLMs, achieving improved accuracy. This research did not consider the role of competitive mechanisms in augmenting agent capabilities. However, it laid the groundwork with valuable data and models to support our subsequent study.

\subsection{Multi-agent Problem Solving}

The primary motivation for using LLM-based multi-agent to solve problems lies in integrating the collective intelligence of multiple agents with specialized knowledge\citep{guo2024largelanguagemodelbased}. Agents, with their dynamic learning and task allocation capabilities, can significantly enhance LLMs' predictive performance \cite{xi2023risepotentiallargelanguage}. These agents collaborate as independent entities, aiming to efficiently tackle complex challenges such as software development\citep{qian2024chatdevcommunicativeagentssoftware,ruan2023tptulargelanguagemodelbased,dong2024selfcollaborationcodegenerationchatgpt}, agent embodiment\citep{mandi2023rocodialecticmultirobotcollaboration,zhang2024buildingcooperativeembodiedagents}, scientific experiments\citep{Zheng2023}, and scientific debates\citep{Xiong_2023, du2023improvingfactualityreasoninglanguage}.  

Research into multi-agent competition is largely centered on social simulation\citep{junprung2023exploringintersectionlargelanguage,wu2024shallteamupexploring,zhao2024competeaiunderstandingcompetitiondynamics}, with scant attention given to the role of competitive mechanisms in bolstering the task-performance abilities of agents, particularly within the context of time series prediction. Existing studies have provided a reference for the design of competitive mechanisms in time series prediction within this research.





\section{Preliminary}
Following previous studies \citep{Wang2024}, the task of news-driven time series forecasting is described as: a time series $X$ (For example, traffic trend) can be segmented into $S$ time series $X_1, X_2, ... X_S$ through the method of sliding time windows for model training. Given a time series $X_s = \{x_1, x_2, …, x_t\}$ in $X$, where $s \leq S$ and $x_i$ ($i \leq t$) is the value at time $i$, the model first collects a set of relevant $d$ news $N_{s} = \{n_1, n_2, …, n_d\}$ from the news database $D$ based on its logic $L$, and then use the selected $d$ news to predict the value $\widetilde{y}_{s,t+1}$ of $X_s$ at time $t+1$.

\section{Methodology}
\begin{figure*}[t]
  \includegraphics[width=\linewidth]{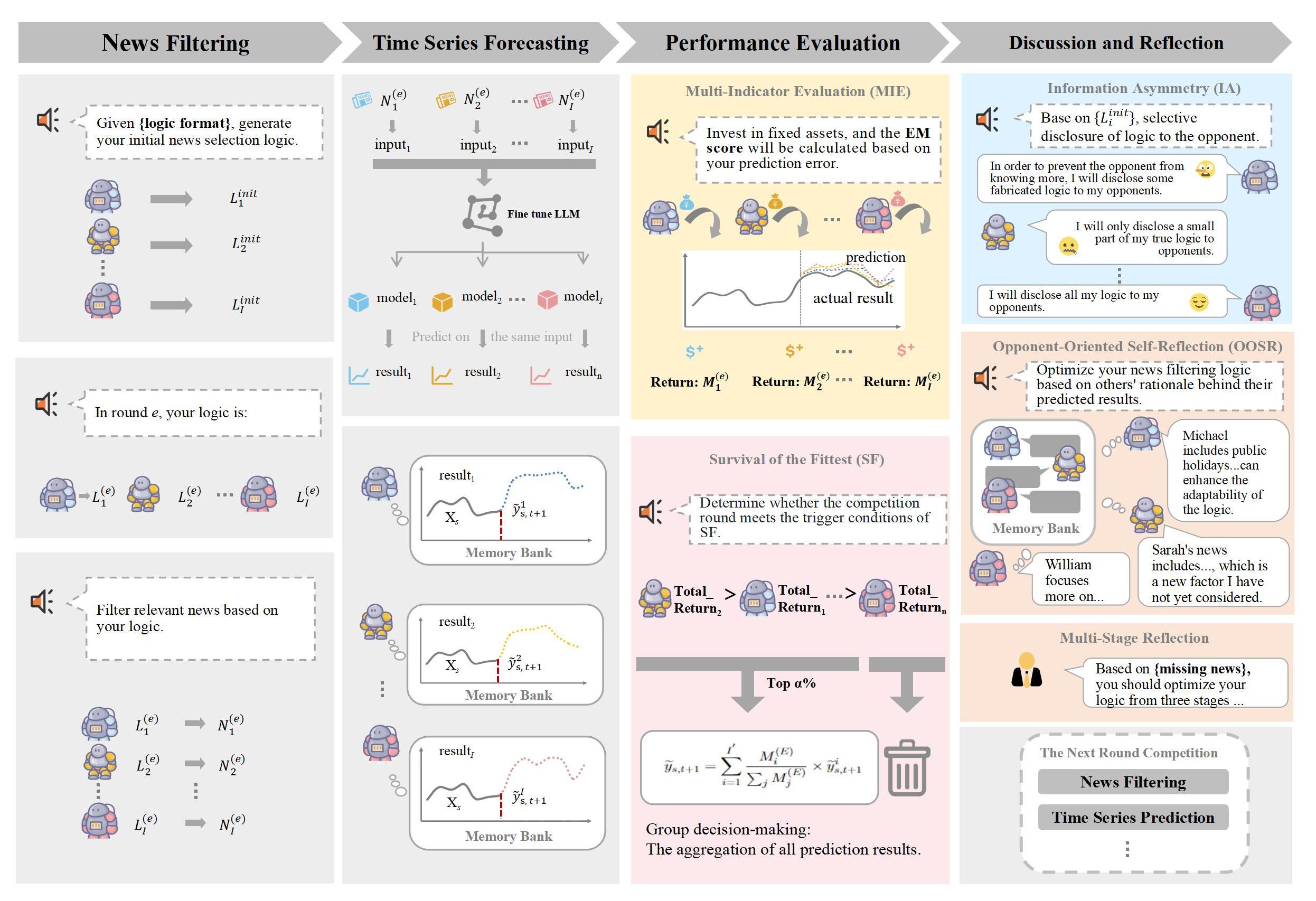}
  \centering
  \caption{The framework of the proposed model.}
  \label{framework}
\end{figure*}

In Figure~\ref{framework}, assume $\text{I}$ agents need to participate in $E$ rounds of competitions. In round $e$, the basic process of the task includes four stages.

\textbf{(1) News Filtering stage:} Each agent $i$ uses its logic $L^{(e)}_{i}$ to select a news set $N^{(e)}_{i} = \{N^{(e)}_{i,1}, N^{(e)}_{i,2}, ..., N^{(e)}_{i,S}\}$ from news database $D$, where $N^{(e)}_{i,s}$ is the selected news set for time series $X_s$ by agent $i$ in round $e$. This process is executed by a LLM named as $\text{LLM}_L$. The prompt template used in this stage mainly draws on the research proposed by \citet{Wang2024}.
    
\textbf{(2) Time Series Forecasting stage:} Each agent $i$ fine-tunes their own LLM model $\text{LLM}^{(e)}_{S,i}$ to predict the value $\widetilde{y}^{i}_{s,t+1}$ at time $t+1$ of each $X_s$ based on the analysis of the selected news $N^{(e)}_{i,s}$. The use of $\text{LLM}^{(e)}_{S,i}$ for training and testing are mainly based on the research proposed by \citet{Wang2024}.
    
\textbf{(3) Agent Performance Evaluation stage:} Each agent’s performance will be evaluated based on evaluation metrics $EM$ by using \textbf{Multi-Indicator Evaluation (MIE)}. The $EM$ score can make agents aware of their own performances in the competition and motivate them to optimize logics. This process is also executed by $\text{LLM}_L$. A \textbf{Survival of Fittest (SF)} is proposed in this stage to eliminate agents with weaker performances.
    
\textbf{(4) Discussion and Reflection stage:} According to the $EM$ score, agents update their logics based on discussion and reflection. An \textbf{Information Asymmetry (IA)} component is proposed in the discussion to allow agents to publish misleading logics and explanations to their opponents. An \textbf{Opponent-Oriented Self-Reflection (OOSR)} is proposed to update agents' logics based on inferring their opponents' logics. The updated logic $L^{(e+1)}_{i}$ of each agent $i$ will be adopted for the $e+1$ round of news selection. This process is also executed by $\text{LLM}_L$. OOSR adopts a \textbf{Multi-stage Reflection (MSR)} strategy to reduce the wrong logic propagation error, which is more prominent in the competitive mechanism. 

Each agent possesses news logic generation, news filtering, time series forecasting, discussion, and reflection capabilities.  Competition is facilitated by the Multi-agent Interactive Environment (MIE), Innovative Agent (IA), Opponent-aware Strategy Optimization and Reflection (OOSR), and Selective Filter (SF) components, which are detailed below.

\subsection{Multi-Indicator Evaluation (MIE)} \label{evaluation}
Drawing on the theory of competition awareness \citep{Chen2025@compet}, we design evaluation metrics (EM) based on each agent's performance on the time series prediction task in formula ~(\ref{eq1}). The performance mainly uses Mean Absolute Percentage Error (MAPE) to evaluate the deviation between true  $y_{t+1}$ and predicted value $\widetilde{y}_{t+1}$.  

\begin{equation} \label{eq1}
\begin{aligned}
\text{EM}^{(e)}_{i} = \{rank^{(e)}_{i}, top^{(e)}_{i}, ave^{(e)}_{i}\}
\end{aligned}
\end{equation}

\noindent where $\text{EM}^{(e)}_{i}$ is the EM score of agent $i$ at round $e$. It contains three indicators: $rank^{(e)}_{i}$ is the ranking of agent $i$ based on its performance in round $e$. $ave^{(e)}_i$ is to evaluate the percentage increase or decrease of $i$'s performance relative to the average performances of all agents. If an agent's performance is better than average, then it will receive a negative $ave$ score. $top^{(e)}_i$ is to evaluate the percentage decrease of $i$'s performance relative to the best performance of all agents.   

The advantage of MIE lies in its ability to allow the agent to intuitively perceive the losses incurred from competition, thereby stimulating the agent to explore more diverse thoughts. We calculate the cumulative score (CS): 

\vspace{-0.5cm}

\begin{equation}
\begin{aligned}
M^{(e+1)}_i = M^{(e)}_i + M^{(e)}_i \times (1 - \text{MMN}(\text{MAPE}^{(e)}_i))
\end{aligned}
\end{equation}

\noindent where $\text{MAPE}^{(e)}_i$ is the MAPE of agent $i$ in round $e$, MMN is Maximum and Minimum Normalization. $M^{(e+1)}_i$ reflects the accumulation of agent $i$'s performance for each round, and can measure the long-term stability of the agent’s performance and serve as an indicator for eliminating agents. 

\subsection{Survival of the Fittest (SF)}
After every $E$ rounds of competitions, The SF is invoked where certain agents are eliminated. The SF is designed to guarantee that the good agents can advance to the ultimate stage of group decision-making. The mechanism primarily adheres to the following principles: Agents ranking in the bottom (1-$\alpha$)\% based on their CS scores will be eliminated, where $\alpha$ determines the retention ratio of agents. 


\subsection{Information Asymmetry (IA)} 
In a discussion, IA embodies information asymmetry from two aspects: First, IA allows an agent to send information to all agents, or choose to send information to selected agents (\textbf{Selective communication}). As introduced in previous study ~\citep{wu2024shallteamupexploring}, agents will spontaneously cooperate in competition. The mode design can assist agents to adopt more flexible strategies to decide competition or collaboration. Second, IA allows an agent to publish incomplete or misleading logic to other opponents (\textbf{Hide or forge logic}). Information asymmetry is an inherent attribute or strategy to prevent opponents from obtaining a player’s key information. Additionally, research has shown that IA can significantly improve the stability and efficiency of the agents' learning process, outperforming independent learning scenarios \citep{WANG201740,tampubolon2021informationasymmetrycompetitivemultiagent}. The output of IA is described as below:

\vspace{-0.5cm}

\begin{equation}
\begin{aligned}
\text{PL}^{(e)} = \{pl^{(e)}_1, pl^{(e)}_2, ..., pl^{(e)}_I\}
\end{aligned}
\end{equation}

\vspace{-0.25cm}

\noindent where $\text{PL}^{(e)}$ is the set of logic, which are published by each agent in round $e$. $pl^{(e)}_i$ is the logic and its explanation published by agent $i$ where $i \leq I$. The detailed description of $pl^{(e)}_i$ can be seen in Appendix A.4.     
 
\subsection{Opponent-Oriented Self-Reflection (OOSR)} 
After IA component, each agent can update its own news selection logic by referencing the logic $\text{PL}^{(e)}$ of others. Due to the presence of incomplete and misleading information in the $\text{PL}^{(e)}$, the wrong logic propagation error will be magnified. We propose the MSR model to enhance agents’ ability to discriminate against misleading logic.

\textbf{Multi-Stage Reflection (MSR). }MSR contains three stages. In the first stage, following the method proposed by \citet{Wang2024@rethinking,Wang2024}, each agent updates its news selection logic as $L^{(e+1)'}_i$.

In the second stage, we design a $\text{diff}$ function to extract the updated parts from $L^{(e+1)'}_i$ compared with $L^{(e)}_i$. The formula could be seen as below:

\begin{equation} \label{eq4}
\begin{aligned}
\delta^{(e+1)}_{i} & = \{\delta_1, \delta_2, ..., \delta_U\} \\
& = \text{diff} (L^{(e+1)'}_i, L^{(e)}_i)
\end{aligned}
\end{equation}

\noindent where $\delta^{(e+1)}_{i}$ is the set of $U$ updated parts of agent $i$ in round $e$. For the $u$th updated part $\delta_u$ ($u \leq U$) in $\delta^{(e+1)}_{i}$, we use the fine-tuned $\text{LLM}^{(e)}_i$ to evaluate whether it is a "good" or "bad" logic by testing it on a set of randomly selected $K$ time series. The main idea is that removing a "good" $\delta_u$ to the logic $L^{(e+1)'}_i$ can decline the performance, while removing the "bad" one can improve the performance. The significance of MSR lies in our use of quantitative indicators to assist LLMs in making judgments about misleading logic (bad one).

In the third stage, We retain all updated parts marked as "good" in $L^{(e+1)'}_i$, and re-evaluate those marked as "bad" in conjunction with temporal trends to finally determine whether to keep them. Reflection in this stage ensures that an excessive number of updated parts is not discarded. Assume the final removed parts are $\delta^{(e+1)}_{i,bad}$, and the final logic $L^{(e+1)}_{i}$ for the next round is expressed as:

\begin{equation}
\begin{aligned}
L^{(e+1)}_{i} = L^{(e+1)'}_{i} - \delta^{(e+1)}_{i,bad}
\end{aligned}
\end{equation}

\noindent where the minus sign indicates removing $\delta^{(e+1)}_{i,bad}$ from $L^{(e+1)'}_{i}$. The detailed description of MSR can be seen in Appendix A.5.

\subsection{Aggregation of All Prediction Results}
After the $E$th round of competitions, $\text{I}^{'}$ agents are retained. For a time series $X_s$, each agent can predict the value $\widetilde{y}^{i}_{s,t+1}$ of $X_s$ at time $t+1$ based on its own logic. The aggregation of all prediction results is expressed as:

\begin{equation}
\begin{aligned}
\widetilde{y}_{s,t+1} = \sum^{I^{'}}_{i=1}\frac{M^{(E)}_{i}}{\sum_j M^{(E)}_{j} } \times \widetilde{y}^{i}_{s,t+1}
\end{aligned}
\end{equation}

\noindent where $M^{(E)}_{i}$ is the CS value of agent $i$ in round $E$. $\widetilde{y}_{s,t+1}$ is the aggregated value. We allow the model to complete one training cycle over all time series with $E$ rounds of competitions (The training data is correspondingly divided into $E$ parts), which are defined as one epoch. The termination condition for model training is that $\widetilde{y}_{s,t+1}$ cannot be closer to the true value, or the model completes the training for the specified number of epochs.

\section{Experiments}

\subsection{Datasets and Experimental Setting}

The time-series datasets and corresponding publicly available news datasets in the experiment are from the research of \citet{Wang2024}. These datasets include traffic volume\citep{kuznetsov2017web}, exchange rates\citep{lai2018modeling}, Bitcoin prices\citep{godahewa2021monash}, and Australian electricity demand\citep{godahewa2021monash}. All time series are divided into training, validation, and testing sets, with a ratio of 8:1:1. The training process consists of 3 epochs, which includes 5 rounds of competition, and the training dataset is correspondingly divided into 5 parts through shuffling. The large-scale $\text{LLM}_L$ is conducted on GPT-4o, and the small-scale $\text{LLM}_S$ is conducted on Llama 7B. The number of agents is set at 10. The $\alpha$ value of SF is set at 0.3. 5-fold is adopted for validation. Detailed descriptions of parameter assignments can be seen in Appendix~\ref{Parameter Sensitivity Analysis} and Appendix~\ref{sec:prompt}.  

The baselines compared in Table\ref{tab:comparison_results} include Autoformer~\citep{wu2021autoformer}, Informer~\citep{zhou2021informer}, DLinear~\citep{zeng2023transformers}, iTransformer~\citep{liu2023itransformer}, Frequency Improved Legendre Memory Model (FiLM)~\citep{zhou2022film}, TimesNet~\citep{wu2022timesnet}, Pyraformer~\citep{liu2021pyraformer}, PatchTST~\citep{nie2022transformers}, Fedformer~\citep{zhou2022fedformer}, and GPT4TS~\citep{zhou2024one}. We also introduce three multi-agent based baselines: The agents discussion (AD) adopts the method proposed by \citet{Liang2024@encourage} to update news selection logic through discussions. The agents collaboration (AC) is mainly based on the research of \citet{Wang2024}. The AVE model calculates the average performance of all agents in our model, primarily assessing the improvement in the capability of individual agent.

\begin{table*}[htbp]
\centering
\setlength{\tabcolsep}{6pt} 
\scriptsize  
\begin{tabular}{llccccccccccc}
\toprule
\textbf{Dataset} & \textbf{Metrics} & \textbf{Ours} & 
\textbf{Auto.} & \textbf{In.} & \textbf{Dlin.} & \textbf{iTrans.} &
\textbf{FiLM} & \textbf{Pyra.} &
\textbf{PatchTST} & \textbf{FED.} & \textbf{GPT4TS} \\

\midrule
\multirow{4}{*}{\textbf{Electricity}}
& MAE & {\color[HTML]{FF0000} \textbf{229.19}} & 349.43 & 282.56 & 255.70 & {\color{blue}\textbf{233.58}} & 254.05  & 544.64  & 234.46 & 238.77 & 236.91 \\
& MSE$_{\times10^{-3}}$ & {\color[HTML]{FF0000} \textbf{132.87}}& 251.79 & 166.07 & 161.59 & 135.27 & 153.90  & 625  & {\color{blue} \textbf{133.53}} & 133.96 & 142.60 \\
& RMSE & {\color{blue}\textbf{364.52}} & 407.52 & 401.98 & 367.49 & {\color[HTML]{FF0000} \textbf{312.42}} & 392.38  & 790.54  & 365.41 & 365.44 & 377.62 \\
& MAPE & {\color[HTML]{FF0000} \textbf{6.71\%}} & 10.63\% & 8.94\% & 7.29\% & 6.86\% & 6.81\%  & 36.26\% & 6.75\% & 34.27\% &{\color{blue}\textbf{6.72\%}}  \\ 

\midrule
\multirow{4}{*}{\textbf{Exchange}}
& MAE$_{\times10^{3}}$ & {\color{blue}\textbf{4.41}} &
  9.27 &
  {\color[HTML]{FF0000} \textbf{1.75}} &
  6.96 &
  27.04 &
  5.24 &
  40.18 &
  25.06 &
  35.19 &
  15.05 \\
& MSE$_{\times10^{4}}$ & {\color[HTML]{FF0000} \textbf{0.37}} &
  1.36 &
  4.76 &
  0.91 &
  11.59 &
  {\color{blue}\textbf{0.77}} &
  24.50 &
  10.23 &
  18.45 &
  4.01 \\
& RMSE$_{\times10^{2}}$ &   {\color[HTML]{FF0000} \textbf{0.61}} &
  1.17 &
  2.18 &
  9.52 &
  3.41 &
  {\color{blue}\textbf{0.875}} &
  4.95 &
  3.20 &
  4.30 &
  2.00 \\
& MAPE & {\color[HTML]{FF0000} \textbf{0.63\%}} &
  1.23\% &
  2.32\% &
  0.92\% &
  3.96\% &
  {\color{blue} \textbf{0.70\%}} &
  5.93\% &
  3.68\% &
  5.17\% &
  1.34\% \\

\midrule
\multirow{4}{*}{\textbf{Traffic}}
& MAE$_{\times10^{2}}$ & {\color[HTML]{FF0000} \textbf{1.56}} &
  2.49 &
  4.44 &
  1.70 &
  {\color[HTML]{FF0000} \textbf{1.56}} &
  {\color{blue}\textbf{1.61}} &
  1.69 &
  1.84 &
  1.74 &
  1.64 \\
& MSE$_{\times10^{4}}$ & {\color{blue}\textbf{1.03}} &
  2.19 &
  {\color[HTML]{212121} 5.27} &
  1.67 &
  1.54 &
  1.49 &
  {\color[HTML]{FF0000} \textbf{0.97}} &
  1.54 &
  1.43 &
  1.45 \\
& RMSE$_{\times10^{2}}$ & {\color{blue}\textbf{3.21}} &
  4.68 &
  7.26 &
  4.09 &
  3.93 &
  3.86 &
  {\color[HTML]{FF0000} \textbf{3.12}} &
  3.92 &
  3.79 &
  3.81  \\

\midrule
\multirow{4}{*}{\textbf{Bitcoin}}
& MAE$_{\times10^{-3}}$ & {\color[HTML]{FF0000} \textbf{0.25}} &
  4.28 &
  {\color[HTML]{212121} 12.27} &
  5.74 &
  3.20 &
  3.17 &
  {\color[HTML]{212121} 9.22} &
  2.85 &
  3.96 &
  {\color{blue}\textbf{2.84}} \\
& MSE$_{\times10^{-6}}$ & {\color[HTML]{FF0000} \textbf{0.14}} &
  27.64 &
  {\color[HTML]{212121} 162.47} &
  50.90 &
  16.21 &
  16.38 &
  {\color[HTML]{212121} 123.71} &
  {\color{blue}\textbf{13.52}} &
  24.60 &
  13.66  \\
& RMSE$_{\times10^{-2}}$ & {\color[HTML]{FF0000} \textbf{3.71}} &
  5.26 &
  {\color[HTML]{212121} 12.75} &
  7.13 &
  4.03 &
  4.05 &
  {\color[HTML]{212121} 11.12} &
  {\color{blue}\textbf{3.68}} &
  4.96 &
   3.70 \\
& MAPE &   {\color[HTML]{FF0000}\textbf{2.83\%}} &
  7.61\% &
  {\color[HTML]{212121} 21.28\%} &
  10.39\% &
  5.70\% &
  5.64\% &
  16.16\% &
  {\color[HTML]{212121} 5.13\%} &
  6.97\% &
  {\color{blue}\textbf{5.08\%}} \\
\bottomrule
\end{tabular}
\caption{Performances on four datasets. Compared with 9 Deep Neural Network based baselines. Elements in red color are the best results, those with blue color are second-best.}
\label{tab:comparison_results}
\end{table*}

\subsection{Metrics}
We consider four metrics, which are commonly used in the corresponding tasks. These are RMSE, MSE, MAE and MAPE \citep{zhou2023ptse,Wang2024}. To effectively evaluate the ability of competitive mechanisms to foster innovative thinking, we utilize bge-m3 \citep{Chen2025@bgem3} to vectorize the logic, with bge-m3 capable of encoding texts up to a maximum of 8192 tokens. We use cosine similarity to measure the similarity $sim(logic_1, logic_2)$ between two logics. The more similar two logics are, the less innovative thought in $logic_2$ is with respect to $logic_1$. If an agent's logic at current epoch is $logic_1$, at epoch + 1 is $logic_2$, we use $1-sim(logic_1, logic_2)$ to measure the \textbf{Logic Update Degree (LUD)} of $logic_2$ relative to $logic_1$.   

\subsection{Main Results}
The experimental results on four datasets are shown in Table~\ref{tab:comparison_results}. Compared with the best baseline, the average improvements are 31.03\%, 36.31\%,2.48\% and 18.14\% in terms of MAE, MSE, RMSE and MAPE. The model shows particularly significant improvements in MSE and RMSE, indicating that the model is effective in reducing variance volatility and detecting sudden changes. The main reason is that agents within the competitive mechanism are able to enhance their novel thinking and judgment abilities, which allows for a better correlation with news events and temporal fluctuations. 

In Table~\ref{tab:simple_avg_results}, the performance of our AVE model also indicates that the competition can improve individual agent's performances significantly. Compared with agents collaboration (AC) baseline, the improvements of the AVE model are 24.93\%, 57.37\%, 40.71\% and 52.69\% on the four metrics. Compared with agents discussion (AD) baseline, the improvements are 6.74\%, 43.55\%, 32.41\% and 31.87\% in terms of the four metrics. This indicates the importance of adding competition into agent discussion framework.      

\begin{table}[htbp]
\centering
\setlength{\tabcolsep}{6pt} 
\scriptsize 
\begin{tabular}{llcccc}
\toprule
\textbf{Dataset} & \textbf{Metrics} & \textbf{Ours} & \textbf{AVE} & \textbf{AD} & \textbf{AC} \\
\midrule
\multirow{4}{*}{\textbf{Electricity}}
& MAE & \textbf{229.19} & \underline{237.71} & 246.61 & 250.71 \\
& MSE$_{\times10^{-3}}$ & \textbf{132.87} & \underline{175.54} & 203.14 & 192.24 \\
& RMSE & \textbf{364.52} & \underline{418.98} & 450.71 & 438.45 \\
& MAPE & \textbf{6.71\%} & 7.62\% & \underline{6.79\%} & 7.60\% \\ 

\midrule
\multirow{4}{*}{\textbf{Exchange}}
& MAE$_{\times10^{3}}$ & \textbf{4.41} & \underline{10.27} & 11.69 & 13.43 \\
& MSE$_{\times10^{4}}$ & \textbf{0.37} & \underline{1.12} & 19.80 & 17.72 \\
& RMSE$_{\times10^{2}}$ & \textbf{0.61} & \underline{3.34} & 4.45 & 4.21 \\
& MAPE & \textbf{0.63\%} & \underline{1.37\%} & 5.61\% & 5.69\% \\

\midrule
\multirow{4}{*}{\textbf{Traffic}}
& MAE$_{\times10^{2}}$ & \textbf{1.56} & \underline{1.63} & 1.72 & 1.84 \\
& MSE$_{\times10^{3}}$ & \textbf{1.03} & \underline{1.47} & 1.58 & 1.81 \\
& RMSE$_{\times10^{2}}$ & \textbf{3.21} & \underline{3.91} & 3.98 & 4.26 \\
\midrule

\multirow{4}{*}{\textbf{Bitcoin}}
& MAE$_{\times10^{-3}}$ & \textbf{0.25} & 0.37 & \underline{0.26} & 0.51 \\
& MSE$_{\times10^{-6}}$ & \textbf{0.14} & 0.24 & \underline{0.15} & 0.33 \\
& RMSE$_{\times10^{-2}}$ & \textbf{3.71} & 4.94 & \underline{3.90} & 5.78 \\
& MAPE & \textbf{2.83\%} & 5.68\% & \underline{3.00\%} & 6.65\% \\

\bottomrule
\end{tabular}
\caption{Comparison of the performances of our model with AVE, AD \citep{Liang2024@encourage} and AC \citep{Wang2024} models, which are multi-agent based framework. Elements in bold are the best results, those with underline are second-best.}
\label{tab:simple_avg_results}
\end{table}

\vspace{-0.5cm}

\subsection{Ablation Study}
To demonstrate the effectiveness of each model component, we compare the complete competition mechanism with 4 variants as follows.

\textbf{CM-IA}: We remove the IA component from the competition mechanism.
\textbf{CM-MIE}: We remove the three evaluation indicators from the competition mechanism.
\textbf{CM-SF}: We remove the Survival of Fittest component from the competition mechanism.
\textbf{CM-MSR}: We remove the 2nd and 3rd stages from MSR component, and only keep the 1st stage, which is based on the method proposed by \citet{Wang2024@rethinking,Wang2024}.
\textbf{CM} is the complete competition mechanism.

\begin{table*}
\centering
\scriptsize
\begin{tabular}{@{}l|cccccccccccc|cccc@{}}
\cmidrule(r){1-17}
\multicolumn{1}{l|}{\multirow{2}{*}{}}      & \multicolumn{12}{c|}{\textbf{Electricity}}                                                                                                               & \multicolumn{4}{c}{\textbf{Exchange}}                                                                                                              \\ \cmidrule(lr){2-17}
\multicolumn{1}{l|}{}                       & \multicolumn{3}{c|}{RMSE}           & \multicolumn{3}{c|}{MSE\( \times 10^{-3} \)} & \multicolumn{3}{c|}{MAE}           &  \multicolumn{3}{c|}{MAPE\( \times 10^{2} \)}          & \multicolumn{1}{c|}{RMSE\( \times 10^{2} \)}          & \multicolumn{1}{c|}{MSE\( \times 10^{4} \)} & \multicolumn{1}{c|}{MAE\( \times 10^{3} \)}          & MAPE\( \times 10^{2} \)          \\ \cmidrule(r){1-17}
CM-IA (Remove IA from CM)            & \multicolumn{3}{c|}{450.71}          & \multicolumn{3}{c|}{203.14}               & \multicolumn{3}{c|}{\underline{246.61}}          & \multicolumn{3}{c|}{\underline{6.79}}               & \multicolumn{1}{c|}{\underline{4.45}}        & \multicolumn{1}{c|}{\underline{19.80}}               & \multicolumn{1}{c|}{\underline{11.69}}       & 5.61               \\
CM-MIE (Remove MIE from CM)           & \multicolumn{3}{c|}{446.88}          & \multicolumn{3}{c|}{199.70}               & \multicolumn{3}{c|}{254.33}          & \multicolumn{3}{c|}{7.75}               & \multicolumn{1}{c|}{4.55}        & \multicolumn{1}{c|}{20.72}               & \multicolumn{1}{c|}{31.72}       & \underline{4.49}               \\
CM-MSR (Replace MSR from CM) & \multicolumn{3}{c|}{443.25}          & \multicolumn{3}{c|}{196.47}               & \multicolumn{3}{c|}{252.55}          & \multicolumn{3}{c|}{7.65}                & \multicolumn{1}{c|}{5.98}        & \multicolumn{1}{c|}{35.88}               & \multicolumn{1}{c|}{46.41}       & 6.60               \\
CM-SF (Remove SF from CM) & \multicolumn{3}{c|}{\underline{439.22}}          & \multicolumn{3}{c|}{\underline{192.92}}               & \multicolumn{3}{c|}{250.58}          & \multicolumn{3}{c|}{7.49}                & \multicolumn{1}{c|}{5.94}        & \multicolumn{1}{c|}{35.31}               & \multicolumn{1}{c|}{45.99}       & 6.53               \\
CM (Complete Competition Mechanism) & \multicolumn{3}{c|}{\textbf{364.52}}          & \multicolumn{3}{c|}{\textbf{132.87}}               & \multicolumn{3}{c|}{\textbf{229.19}}          & \multicolumn{3}{c|}{\textbf{6.71}}              & \multicolumn{1}{c|}{\textbf{0.61}}        & \multicolumn{1}{c|}{\textbf{0.37}}               & \multicolumn{1}{c|}{\textbf{4.41}}       & \textbf{0.63}               \\
\cmidrule(r){1-17}
\multirow{2}{*}{}                           & \multicolumn{12}{c|}{\textbf{Traffic}}                                                                                                                   & \multicolumn{4}{c}{\textbf{Bitcoin}}                                                                                                               \\ \cmidrule(lr){2-17}
                                            \multicolumn{1}{l|}{}                       & \multicolumn{4}{c|}{RMSE\(\times 10^{2} \)}           & \multicolumn{4}{c|}{MSE\( \times 10^{3} \)} & \multicolumn{4}{c|}{MAE\( \times 10^{2} \)}           & \multicolumn{1}{c|}{RMSE\( \times 10^{-2} \)}          & \multicolumn{1}{c|}{MSE\( \times 10^{-6} \)} & \multicolumn{1}{c|}{MAE\( \times 10^{-3} \)}          & MAPE\( \times 10^{2} \)          \\ \cmidrule(r){1-17}
CM-IA (Remove IA from CM)            & \multicolumn{4}{c|}{\underline{3.98}}          & \multicolumn{4}{c|}{\underline{1.58}}               & \multicolumn{4}{c|}{\underline{1.72}}                       & \multicolumn{1}{c|}{\underline{3.90}}        & \multicolumn{1}{c|}{\underline{0.15}}               & \multicolumn{1}{c|}{\underline{0.26}}       & \underline{3.00}            \\

CM-MIE (Remove MIE from CM) & \multicolumn{4}{c|}{10.61}          & \multicolumn{4}{c|}{11.26}               & \multicolumn{4}{c|}{6.25}                      & 


\multicolumn{1}{c|}{3.94}        & \multicolumn{1}{c|}{\underline{0.15}}               & \multicolumn{1}{c|}{\underline{0.26}}       & 3.06             \\
CM-MSR (Replace MSR from CM) & \multicolumn{4}{c|}{7.55}          & \multicolumn{4}{c|}{5.70}               & \multicolumn{4}{c|}{5.19}                    & \multicolumn{1}{c|}{4.16}        & \multicolumn{1}{c|}{0.17}               & \multicolumn{1}{c|}{0.28}       & 3.14             \\
CM-SF (Remove SF from CM)          & \multicolumn{4}{c|}{8.66}          & \multicolumn{4}{c|}{7.50}               & \multicolumn{4}{c|}{5.65}                  & \multicolumn{1}{c|}{4.68}        & \multicolumn{1}{c|}{0.22}               & \multicolumn{1}{c|}{0.31}       & 3.48              \\
CM (Complete Competition Mechanism)              & \multicolumn{4}{c|}{\textbf{3.21}}          & \multicolumn{4}{c|}{\textbf{1.03}}               & \multicolumn{4}{c|}{\textbf{1.56}}        & \multicolumn{1}{c|}{\textbf{3.71}}        & \multicolumn{1}{c|}{\textbf{0.14}}               & \multicolumn{1}{c|}{\textbf{0.25}}       & \textbf{2.83}
               \\
\cmidrule(r){1-17}
\end{tabular}
  \caption{Ablation Study. Elements in bold are the best results, those with underline are second-best.}
  \label{tab:Ablation}
\end{table*}

Ablation studies (Table ~\ref{tab:Ablation}) show the importance of key components. Removing the Innovative Agent (IA) significantly hurts performance, highlighting its role in innovation and robustness. Eliminating the Multi-agent Interactive Environment (MIE) weakens competitive awareness. Removing the Selective Filter (SF) causes a 20.49\% performance drop, emphasizing its contribution to quality. Finally, replacing Multi-Stage Reflection (MSR) with traditional discussion significantly reduces performance, validating the use of fine-tuned LLMs for enhanced decision-making. This also indicates that the introduction of MSR significantly improves agents’ ability to identify misleading information, thereby enabling them to more accurately select news for prediction.

\subsection{Effectiveness of IA for Creating Novel Thought}

\begin{figure}[!ht]
  \includegraphics[width=\columnwidth]{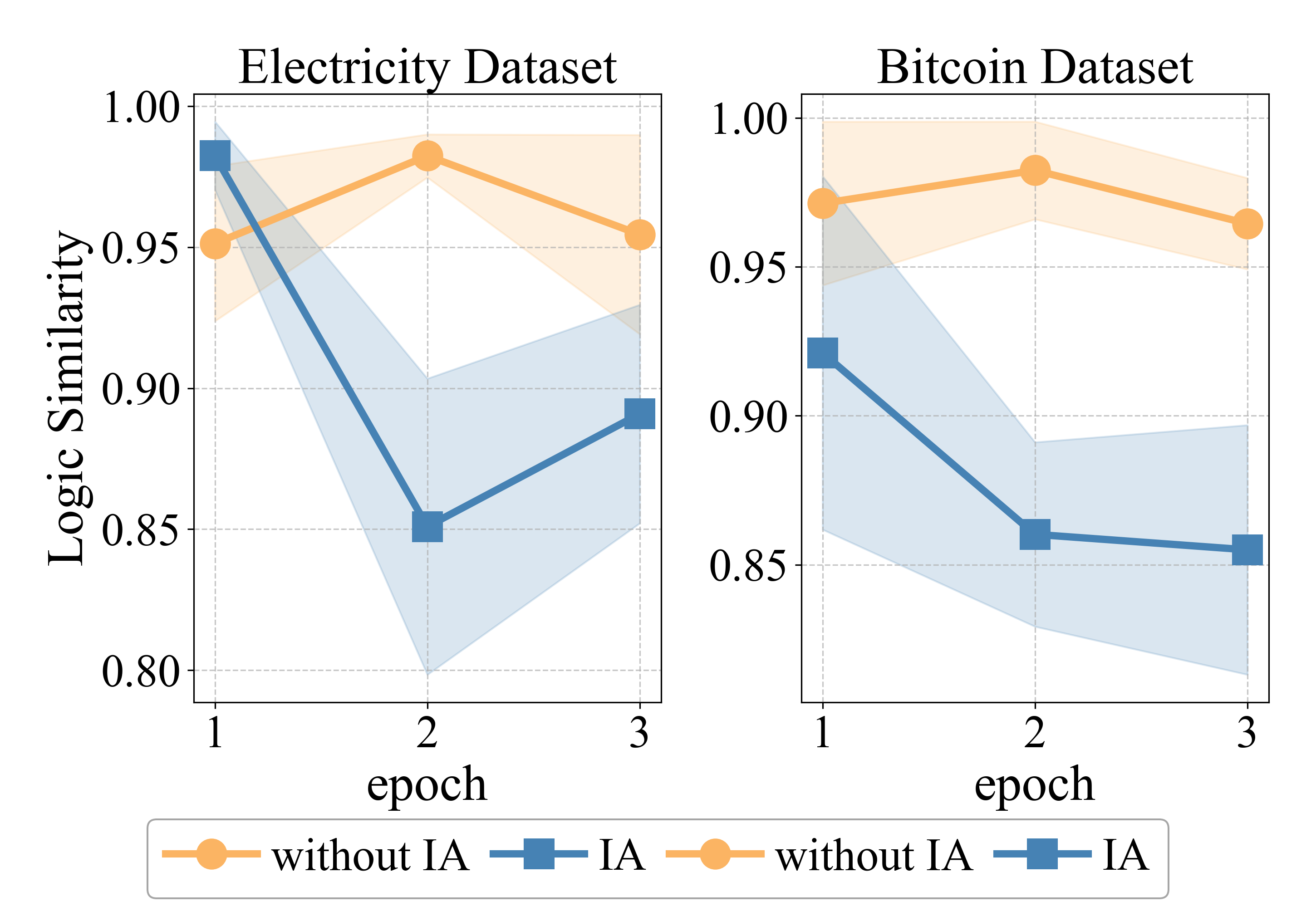} 
  \caption{
This section compares logic similarity between models with and without Information Asymmetry (IA), using electricity and Bitcoin datasets. Higher logic similarity indicates less innovative thinking.}
  \label{fig:similarity}
\end{figure}

Figure~\ref{fig:similarity} shows the average logical similarity of all agents at the end of each epoch. Compared to the model without IA (blue line), the IA component helps maintain lower logical similarity, indicating more diversified agent logics. This is because IA enables agents to conceal or fabricate information, reducing groupthink and promoting diverse strategies, which encourages agents to explore and validate more innovative ideas.

\begin{figure}[!ht]
  \includegraphics[width=\columnwidth]{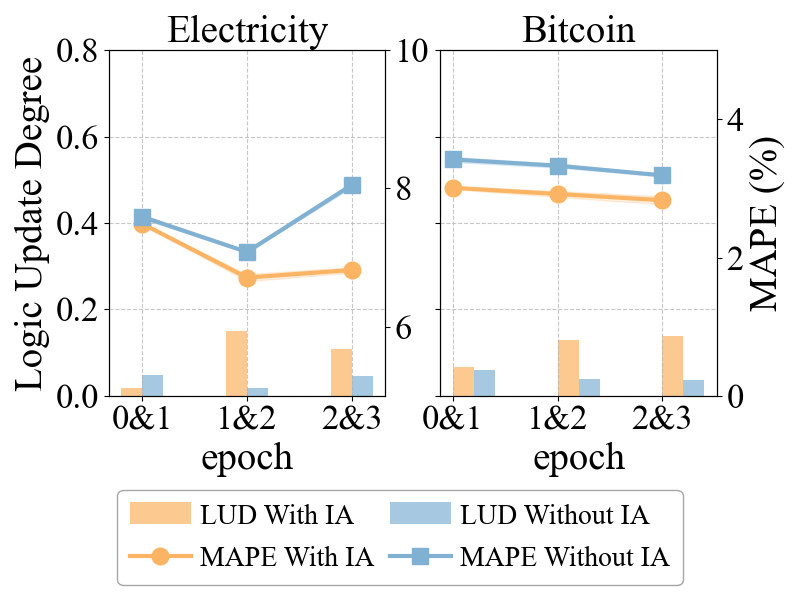} 
  \caption{
This figure compares logic update degree and MAPE across epochs for models with and without IA, using electricity and Bitcoin datasets.
  }
  \label{fig:logic_update_degree}
\end{figure}

Figure~\ref{fig:logic_update_degree} evaluates the degree of logical update for each agent between adjacent epochs. Compared with models without IA (blue line), the model with IA (red line) enables agents to produce more significant logic updates after multiple rounds of competition. This demonstrates that IA prompts agents to generate novel thoughts and continuously update their logic. The MAPE comparison further reinforces the idea that enhancing innovative thinking can significantly improve agents' performance.

\subsection{Effectiveness of MIE for Creating Novel Thought}
\begin{figure}[!ht]
  \includegraphics[width=\columnwidth]{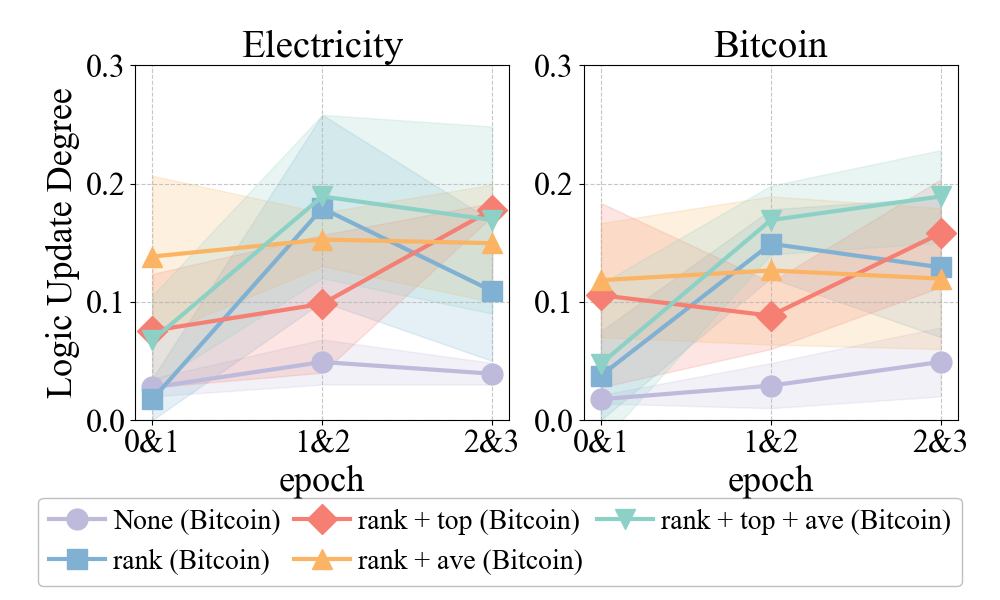} 
  \caption{
Comparison of Logic Update Degree ($rank$, $top$, $ave$) across three epochs in IA and no IA contexts, demonstrating the impact of competition on innovative thinking. }
  \label{fig:Rank&Gap}
\end{figure}

This experiment aims to ascertain whether agents can be motivated to engage in innovative thinking by recognizing their own position within the competitive group, as indicated by the $rank$, $top$ and $ave$ indicators. Figure~\ref{fig:Rank&Gap} evaluates the role of the three indicators on agent’s competition awareness. Compared with not using any of the three indicators (None), considering $rank$, $rank+ave$, and $rank+top$ all have impacts on the degree to which the agent’s logic is updated. The settings are detailed in Prompt 16 to 18 in Appendix~\ref{prompt}.

The model that includes all indicators allows agents of different rankings to fully perceive their respective status, thus exhibiting a higher degree of logical update in the first and second epochs. In the third epoch, due to the elimination of some agents, the LUD of the remaining high-level agents decreased. Additionally, the fluctuation range of the value domain for each indicator is significant, indicating that there is a large variation in the degree of logic update among different agents.

\subsection{The Relationship between Competition Intensity and Model Performance}

In this experiment, we discuss whether the competitive intensity of an agent will have an impact on its performance. We define competitive degree (CPD) based on the calculation of collaborative degree (CLD). The detailed definition of CLD and CPD can be see in Appendix A.6.  

\begin{figure}[!ht]
  \includegraphics[width=\columnwidth]{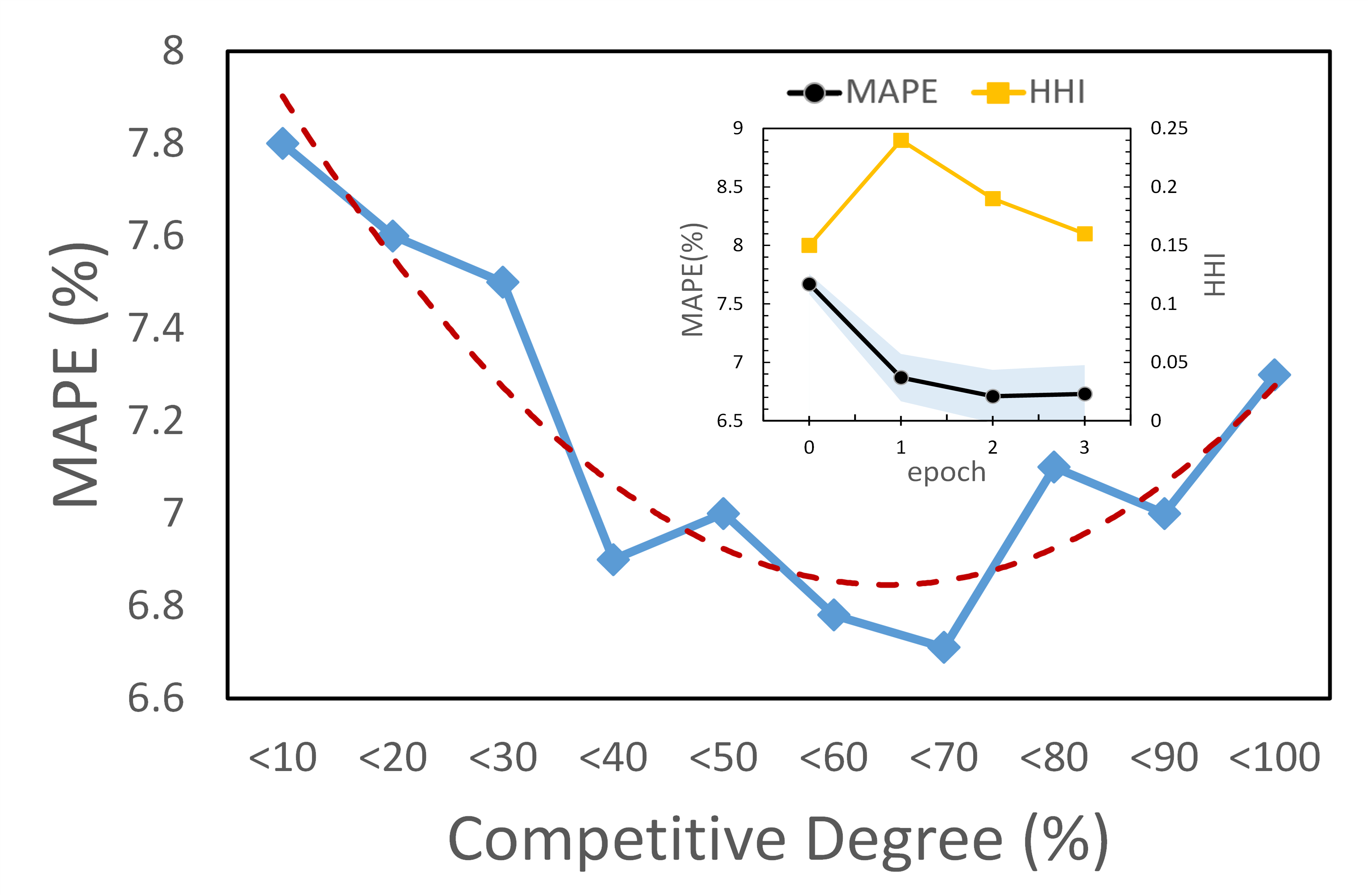} 
  \caption{
This figure shows the relationship between MAPE and the competitive degrees of different agents. The U-shaped trend indicates that MAPE gets its optimal value when when competition is at a moderate level. HHI is the Herfindahl-Hirschman Index.
  }
  \label{fig:Comp&Perf}
\end{figure}

As indicated in the experimental settings, each agent's performance in each round will be evaluated, their competitive degrees will also be calculated. Herfindahl-Hirschman Index ~\citep{Herfindahl1950Concentration} is also adopted to calculate the competitive intensity through the CS scores of each agent. In the main plot of Figure~\ref{fig:Comp&Perf}, we can observe that agents with a moderate level of competitive degree generally perform better on average. Their degrees are mainly concentrated between 40\%-70\%, indicating that their strategies have formed a certain balance between competition and cooperation. This will form a U-shaped relationship between competition degree and model performance, as indicated by the red dashed line in the Figure. This finding is consistent with existing sociological research. The trend of the HHI in the subplot shows that the competitive intensity first increases and then decreases, suggesting the presence of spontaneous cooperation, which can enhance the model’s performance (as indicated by the MAPE line in the subplot). The small fluctuation range of the MAPE line in the subplot suggests that the randomness of the LLM has little interference on this model.    

\section{Conclusions}
In this paper, we introduce a competition mechanism to enhance the performance of agents on news-driven time series forecasting. We integrate Information Asymmetry, Competition Awareness, and Survival of the Fittest within the multi-agent discussion framework to stimulate the innovative thinking of agents. Additionally, we introduce MSR to enhance the model’s ability to discriminate against misleading logic. Experimental findings indicate that the competition mechanism effectively bolsters the agents’ capabilities in both innovative thinking and the identification of error logics. 

\section*{Limitations}

Although our paper demonstrates the positive effects of information asymmetry and perception of competition in competitive mechanisms in fostering innovative thinking among agents, the underlying mechanisms still require further investigation. Only by enhancing the controllability of the process can we increase the value of this research. Our subsequent plan involves theoretically exploring and innovating these mechanisms by integrating distillation with chain-of-thought fine-tuning. Additionally, current model predictions seldom consider the integration of mathematical knowledge related to multivariate time series, such as deep auto-regressive time series modeling, time stationary analysis, co-integration testing, and DTW algorithms. This knowledge is crucial for the model to deeply understand the mechanisms behind news and temporal fluctuations and represents a key research direction we will focus on in the future. Lastly, the current multi-agent competitive model demands high computational resources and long computation times, and it faces numerous limitations in complex reasoning tasks over long texts. Moving forward, we will conduct research on optimizing computational resources and operational efficiency.

\bibliography{custom}

\begin{thebibliography}{55}
\providecommand{\natexlab}[1]{#1}

\bibitem[{Balepur et~al.(2024)Balepur, Palta, and Rudinger}]{Balepur2024@llmer}
Nishant Balepur, Shramay Palta, and Rachel Rudinger. 2024.
\newblock It’s not easy being wrong: Large language models struggle with process of elimination reasoning.
\newblock In \emph{Findings of the Association for Computational Linguistics: ACL 2024}, pages 10143--10166.

\bibitem[{Cao et~al.(2023)Cao, Jia, Arik, Pfister, Zheng, Ye, and Liu}]{cao2023tempo}
Defu Cao, Furong Jia, Sercan~O Arik, Tomas Pfister, Yixiang Zheng, Wen Ye, and Yan Liu. 2023.
\newblock Tempo: Prompt-based generative pre-trained transformer for time series forecasting.
\newblock \emph{arXiv preprint arXiv:2310.04948}.

\bibitem[{Chen et~al.(2024)Chen, Xiao, Zhang, Luo, Lian, and Liu}]{Chen2025@bgem3}
Jianlv Chen, Shitao Xiao, Peitian Zhang, Kun Luo, Defu Lian, and Zheng Liu. 2024.
\newblock \href {https://arxiv.org/abs/2402.03216} {Bge m3-embedding: Multi-lingual, multi-functionality, multi-granularity text embeddings through self-knowledge distillation}.
\newblock \emph{Preprint}, arXiv:2402.03216.

\bibitem[{Chen et~al.(2007)Chen, SU, and Tsai}]{Chen2025@compet}
Mingjer Chen, KuoHsien SU, and Wenpin Tsai. 2007.
\newblock Competitive tension: The awareness-motivation-capability perspective.
\newblock \emph{Academy of Management Journal}, 50.

\bibitem[{Cheng and Chin(2024)}]{Cheng2024sociodojo}
Junyan Cheng and Peter Chin. 2024.
\newblock Sociodojo: Building lifelong analytical agents with real-world text and time series.
\newblock In \emph{ICLR'24}, pages 1--35.

\bibitem[{Das et~al.(2023)Das, Kong, Sen, and Zhou}]{das2023decoder}
Ankan Das, Wenzheng Kong, Rahul Sen, and Yan Zhou. 2023.
\newblock A decoder-only foundation model for time-series forecasting.
\newblock \emph{arXiv preprint arXiv:2310.10688}.

\bibitem[{Dong et~al.(2024)Dong, Jiang, Jin, and Li}]{dong2024selfcollaborationcodegenerationchatgpt}
Yihong Dong, Xue Jiang, Zhi Jin, and Ge~Li. 2024.
\newblock Self-collaboration code generation via chatgpt.

\bibitem[{Du et~al.(2023)Du, Li, Torralba, Tenenbaum, and Mordatch}]{du2023improvingfactualityreasoninglanguage}
Yilun Du, Shuang Li, Antonio Torralba, Joshua~B. Tenenbaum, and Igor Mordatch. 2023.
\newblock \href {https://arxiv.org/abs/2305.14325} {Improving factuality and reasoning in language models through multiagent debate}.
\newblock \emph{Preprint}, arXiv:2305.14325.

\bibitem[{Gao et~al.(2024)Gao, Li, Pan, Kuang, Ma, Qian, Wei, Zhang, Xie, Chen, Yao, Peng, Zhang, Zhu, Cheng, Shi, Li, Ding, and Zhou}]{agentscope}
Dawei Gao, Zitao Li, Xuchen Pan, Weirui Kuang, Zhijian Ma, Bingchen Qian, Fei Wei, Wenhao Zhang, Yuexiang Xie, Daoyuan Chen, Liuyi Yao, Hongyi Peng, Ze~Yu Zhang, Lin Zhu, Chen Cheng, Hongzhu Shi, Yaliang Li, Bolin Ding, and Jingren Zhou. 2024.
\newblock Agentscope: A flexible yet robust multi-agent platform.
\newblock \emph{CoRR}, abs/2402.14034.

\bibitem[{Garza and Mergenthaler-Canseco(2023)}]{garza2023timegpt}
Aldo Garza and Martin Mergenthaler-Canseco. 2023.
\newblock Timegpt-1.
\newblock \emph{arXiv preprint arXiv:2310.03589}.

\bibitem[{Godahewa et~al.(2021)Godahewa, Bergmeir, Webb, Hyndman, and Montero-Manso}]{godahewa2021monash}
Rakshitha Godahewa, Christoph Bergmeir, Geoffrey~I Webb, Rob~J Hyndman, and Pablo Montero-Manso. 2021.
\newblock Monash time series forecasting archive.
\newblock \emph{arXiv preprint arXiv:2105.06643}.

\bibitem[{Gruver et~al.(2024)Gruver, Finzi, Qiu, and Wilson}]{gruver2024large}
Nate Gruver, Marc Finzi, Shikai Qiu, and Andrew~G Wilson. 2024.
\newblock Large language models are zero-shot time series forecasters.
\newblock In \emph{Advances in Neural Information Processing Systems}, volume~36.

\bibitem[{Guan et~al.(2025)Guan, Peng, Hou, and Li}]{Guan@2025mmdere}
Yong Guan, Hao Peng, Lei Hou, and Juanzi Li. 2025.
\newblock Mmd-ere:multi-agent multi-sided debate for event relation extraction.
\newblock In \emph{In Proceedings of the 31th International Conference on Computational Linguistics}, pages 6889--6896.

\bibitem[{Guo et~al.(2024)Guo, Chen, Wang, Chang, Pei, Chawla, Wiest, and Zhang}]{guo2024largelanguagemodelbased}
Taicheng Guo, Xiuying Chen, Yaqi Wang, Ruidi Chang, Shichao Pei, Nitesh~V. Chawla, Olaf Wiest, and Xiangliang Zhang. 2024.
\newblock \href {https://arxiv.org/abs/2402.01680} {Large language model based multi-agents: A survey of progress and challenges}.
\newblock \emph{Preprint}, arXiv:2402.01680.

\bibitem[{Herfindahl and Clemens(1950)}]{Herfindahl1950Concentration}
Herfindahl and Orris C.~Orris Clemens. 1950.
\newblock Concentration in the steel industry.
\newblock \emph{columbia university}.

\bibitem[{Huang et~al.(2025)Huang, Yu, Ma, Zhong, Feng, Wang, Chen, Peng, Feng, Qin, and Liu}]{Huang2025@survey}
Lei Huang, Weijiang Yu, Weitao Ma, Weihong Zhong, Zhangyin Feng, Haotian Wang, Qianglong Chen, Weihua Peng, Xiaocheng Feng, Bing Qin, and Ting Liu. 2025.
\newblock \href {https://doi.org/10.1145/3703155} {A survey on hallucination in large language models: Principles, taxonomy, challenges, and open questions}.
\newblock \emph{ACM Trans. Inf. Syst.}, 43(2).

\bibitem[{Jiang et~al.(2023)Jiang, Sablayrolles, Mensch, Bamford, Chaplot, Casas, Bressand, Lengyel, Lample, Saulnier et~al.}]{jiang2023mistral}
Albert~Q Jiang, Alexandre Sablayrolles, Arthur Mensch, Chris Bamford, Devendra~Singh Chaplot, Diego de~las Casas, Florian Bressand, Gianna Lengyel, Guillaume Lample, Lucile Saulnier, et~al. 2023.
\newblock Mistral 7b.
\newblock \emph{arXiv preprint arXiv:2310.06825}.

\bibitem[{Jin et~al.(2023)Jin, Wang, Ma, Chu, Zhang, Shi, Chen, Liang, Li, Pan et~al.}]{jin2023timellm}
Ming Jin, Shiyu Wang, Lintao Ma, Zhixuan Chu, James~Y Zhang, Xiaoming Shi, Pin-Yu Chen, Yuxuan Liang, Yuan-Fang Li, Shirui Pan, et~al. 2023.
\newblock Time-llm: Time series forecasting by reprogramming large language models.
\newblock \emph{arXiv preprint arXiv:2310.01728}.

\bibitem[{Junprung(2023)}]{junprung2023exploringintersectionlargelanguage}
Edward Junprung. 2023.
\newblock \href {https://arxiv.org/abs/2308.07411} {Exploring the intersection of large language models and agent-based modeling via prompt engineering}.
\newblock \emph{Preprint}, arXiv:2308.07411.

\bibitem[{Kuznetsov et~al.(2017)Kuznetsov, Cukierski, and Maggie}]{kuznetsov2017web}
Vitaly Kuznetsov, Will Cukierski, and Oren~Anava Maggie. 2017.
\newblock Web traffic time series forecasting.

\bibitem[{Lai et~al.(2018)Lai, Chang, Yang, and Liu}]{lai2018modeling}
Guokun Lai, Wei-Cheng Chang, Yiming Yang, and Hanxiao Liu. 2018.
\newblock Modeling long-and short-term temporal patterns with deep neural networks.
\newblock In \emph{The 41st International ACM SIGIR Conference on Research \& Development in Information Retrieval}, pages 95--104.

\bibitem[{Lan et~al.(2024)Lan, Hu, Wang, Wang, Ye, Zhao, Lim, Xiong, and Wang}]{lan2024llmbasedagentsocietyinvestigation}
Yihuai Lan, Zhiqiang Hu, Lei Wang, Yang Wang, Deheng Ye, Peilin Zhao, Ee-Peng Lim, Hui Xiong, and Hao Wang. 2024.
\newblock \href {https://arxiv.org/abs/2310.14985} {Llm-based agent society investigation: Collaboration and confrontation in avalon gameplay}.
\newblock \emph{Preprint}, arXiv:2310.14985.

\bibitem[{Liang et~al.(2024)Liang, He, Jiao, Wang, Wang, Wang, Yang, Shi, and Tu}]{Liang2024@encourage}
Tian Liang, Zhiwei He, Wenxiang Jiao, Xing Wang, Yan Wang, Rui Wang, Yujiu Yang, Shuming Shi, and Zhaopeng Tu. 2024.
\newblock \href {https://arxiv.org/abs/2305.19118} {Encouraging divergent thinking in large models through multi-agent debate}.
\newblock \emph{Preprint}, arXiv:2305.19118.

\bibitem[{Liu et~al.(2021)Liu, Yu, Liao, Li, Lin, Liu, and Dustdar}]{liu2021pyraformer}
Shizhan Liu, Hang Yu, Cong Liao, Jianguo Li, Weiyao Lin, Alex~X. Liu, and Schahram Dustdar. 2021.
\newblock Pyraformer: Low-complexity pyramidal attention for long-range time series modeling and forecasting.
\newblock In \emph{International Conference on Learning Representations}.

\bibitem[{Liu et~al.(2023)Liu, Hu, Zhang, Wu, Wang, Ma, and Long}]{liu2023itransformer}
Yong Liu, Tengge Hu, Haoran Zhang, Haixu Wu, Shiyu Wang, Lintao Ma, and Mingsheng Long. 2023.
\newblock itransformer: Inverted transformers are effective for time series forecasting.
\newblock \emph{arXiv preprint arXiv:2310.06625}.

\bibitem[{Mandi et~al.(2024)Mandi, Jain, and Song}]{mandi2023rocodialecticmultirobotcollaboration}
Zhao Mandi, Shreeya Jain, and Shuran Song. 2024.
\newblock Roco: Dialectic multi-robot collaboration with large language models.

\bibitem[{Nie et~al.(2022)Nie, Nguyen, Sinthong, and Kalagnanam}]{nie2022transformers}
Yuqi Nie, Nam~H. Nguyen, Phanwadee Sinthong, and Jayant Kalagnanam. 2022.
\newblock A time series is worth 64 words: Long-term forecasting with transformers.
\newblock \emph{arXiv preprint arXiv:2211.14730}.

\bibitem[{Qian et~al.(2024)Qian, Liu, Liu, Chen, Dang, Li, Yang, Chen, Su, Cong et~al.}]{qian2024chatdevcommunicativeagentssoftware}
Chen Qian, Wei Liu, Hongzhang Liu, Nuo Chen, Yufan Dang, Jiahao Li, Cheng Yang, Weize Chen, Yusheng Su, Xin Cong, et~al. 2024.
\newblock Chatdev: Communicative agents for software development.

\bibitem[{Rasul et~al.(2023)Rasul, Ashok, Williams, Khorasani, Adamopoulos, Bhagwatkar, Bilo{\v{s}}, Ghonia, Hassen, Schneider, Garg, Drouin, Chapados, Nevmyvaka, and Rish}]{rasul2023lagllama}
Kashif Rasul, Arjun Ashok, Andrew~Robert Williams, Arian Khorasani, George Adamopoulos, Rishika Bhagwatkar, Marin Bilo{\v{s}}, Hena Ghonia, Nadhir Hassen, Anderson Schneider, Sahil Garg, Alexandre Drouin, Nicolas Chapados, Yuriy Nevmyvaka, and Irina Rish. 2023.
\newblock \href {https://openreview.net/forum?id=jYluzCLFDM} {Lag-llama: Towards foundation models for time series forecasting}.

\bibitem[{Rodrigues et~al.(2019)Rodrigues, Markou, and Pereira}]{Rodrigues2019combinetsandtext}
Filipe Rodrigues, Ioulia Markou, and Francisco Pereira. 2019.
\newblock Combining time-series and textual data for taxi demand prediction in event areas: A deep learning approach.
\newblock \emph{Information Fusion}, 49:120--129.

\bibitem[{Ruan et~al.(2023)Ruan, Chen, Zhang, Xu, Bao, Du, Shi, Mao, Li, Zeng, and Zhao}]{ruan2023tptulargelanguagemodelbased}
Jingqing Ruan, Yihong Chen, Bin Zhang, Zhiwei Xu, Tianpeng Bao, Guoqing Du, Shiwei Shi, Hangyu Mao, Ziyue Li, Xingyu Zeng, and Rui Zhao. 2023.
\newblock \href {https://arxiv.org/abs/2308.03427} {Tptu: Large language model-based ai agents for task planning and tool usage}.
\newblock \emph{Preprint}, arXiv:2308.03427.

\bibitem[{Sun et~al.(2023)Sun, Li, Li, and Hong}]{sun2023test}
Chao Sun, Yue Li, Hanyu Li, and Shaopeng Hong. 2023.
\newblock Test: Text prototype aligned embedding to activate llm's ability for time series.
\newblock \emph{arXiv preprint arXiv:2308.08241}.

\bibitem[{Tampubolon et~al.(2021)Tampubolon, Ceribasic, and Boche}]{tampubolon2021informationasymmetrycompetitivemultiagent}
Ezra Tampubolon, Haris Ceribasic, and Holger Boche. 2021.
\newblock \href {https://arxiv.org/abs/2010.10901} {On information asymmetry in competitive multi-agent reinforcement learning: Convergence and optimality}.
\newblock \emph{Preprint}, arXiv:2010.10901.

\bibitem[{Tang et~al.(2025)Tang, Zhang, Jin, Yu, Wang, Jin, Zhang, and Du}]{tang2024timeseriesforecastingllms}
Hua Tang, Chong Zhang, Mingyu Jin, Qinkai Yu, Zhenting Wang, Xiaobo Jin, Yongfeng Zhang, and Mengnan Du. 2025.
\newblock Time series forecasting with llms: Understanding and enhancing model capabilities.

\bibitem[{Team et~al.(2023)Team, Anil, Borgeaud, Alayrac, Yu, Soricut, Schalkwyk, Dai, Hauth, Millican et~al.}]{team2023gemini}
Gemini Team, Rohan Anil, Sebastian Borgeaud, Jean-Baptiste Alayrac, Jiahui Yu, Radu Soricut, Johan Schalkwyk, Andrew~M Dai, Anja Hauth, Katie Millican, et~al. 2023.
\newblock Gemini: a family of highly capable multimodal models.
\newblock \emph{arXiv preprint arXiv:2312.11805}.

\bibitem[{Touvron et~al.(2023)Touvron, Martin, Stone, Albert, Almahairi, Babaei, Bashlykov, Batra, Bhargava, Bhosale et~al.}]{touvron2023llama}
Hugo Touvron, Louis Martin, Kevin Stone, Peter Albert, Amjad Almahairi, Yasmine Babaei, Nikolay Bashlykov, Soumya Batra, Prajjwal Bhargava, Shruti Bhosale, et~al. 2023.
\newblock Llama 2: Open foundation and fine-tuned chat models.
\newblock \emph{arXiv preprint arXiv:2307.09288}.

\bibitem[{Wang and Wang(2017)}]{WANG201740}
Kun~Tracy Wang and Wanbin~Walter Wang. 2017.
\newblock \href {https://doi.org/10.1016/j.econmod.2016.11.024} {Competition in the stock market with asymmetric information}.
\newblock \emph{Economic Modelling}, 61:40--49.

\bibitem[{Wang et~al.(2024{\natexlab{a}})Wang, Wang, Su, Tong, and Song}]{Wang2024@rethinking}
Qineng Wang, Zihao Wang, Ying Su, Hanghang Tong, and Yangqiu Song. 2024{\natexlab{a}}.
\newblock \href {https://arxiv.org/abs/2402.18272} {Rethinking the bounds of llm reasoning: Are multi-agent discussions the key?}
\newblock \emph{Preprint}, arXiv:2402.18272.

\bibitem[{Wang et~al.(2024{\natexlab{b}})Wang, Feng, Qiu, Gu, and Zhao}]{Wang2024}
Xinlei Wang, Maike Feng, Jing Qiu, Jinjin Gu, and Junhua Zhao. 2024{\natexlab{b}}.
\newblock From news to forecast: Iterative event reasoning in llm-based time series forecasting.
\newblock In \emph{Proceedings of the 38th Conference on Neural Information Processing Systems (NeurIPS 2024)}.

\bibitem[{Wu et~al.(2022)Wu, Hu, Liu, Zhou, Wang, and Long}]{wu2022timesnet}
Haixu Wu, Tengge Hu, Yong Liu, Hang Zhou, Jianmin Wang, and Mingsheng Long. 2022.
\newblock Timesnet: Temporal 2d-variation modeling for general time series analysis.
\newblock In \emph{The Eleventh International Conference on Learning Representations}.

\bibitem[{Wu et~al.(2021)Wu, Xu, Wang, and Long}]{wu2021autoformer}
Haixu Wu, Jiehui Xu, Jianmin Wang, and Mingsheng Long. 2021.
\newblock Autoformer: Decomposition transformers with auto-correlation for long-term series forecasting.
\newblock \emph{Advances in Neural Information Processing Systems}, 34:22419--22430.

\bibitem[{Wu et~al.(2024)Wu, Peng, Zheng, Liu, Han, Kwon, Onizuka, Tang, and Xiao}]{wu2024shallteamupexploring}
Zengqing Wu, Run Peng, Shuyuan Zheng, Qianying Liu, Xu~Han, Brian~Inhyuk Kwon, Makoto Onizuka, Shaojie Tang, and Chuan Xiao. 2024.
\newblock \href {https://arxiv.org/abs/2402.12327} {Shall we team up: Exploring spontaneous cooperation of competing llm agents}.
\newblock \emph{Preprint}, arXiv:2402.12327.

\bibitem[{Xi et~al.(2025)Xi, Chen, Guo, He, Ding, Hong, Zhang, Wang, Jin, Zhou et~al.}]{xi2023risepotentiallargelanguage}
Zhiheng Xi, Wenxiang Chen, Xin Guo, Wei He, Yiwen Ding, Boyang Hong, Ming Zhang, Junzhe Wang, Senjie Jin, Enyu Zhou, et~al. 2025.
\newblock The rise and potential of large language model based agents: A survey.

\bibitem[{Xiong et~al.(2023)Xiong, Ding, Cao, Liu, and Qin}]{Xiong_2023}
Kai Xiong, Xiao Ding, Yixin Cao, Ting Liu, and Bing Qin. 2023.
\newblock \href {https://doi.org/10.18653/v1/2023.findings-emnlp.508} {Examining inter-consistency of large language models collaboration: An in-depth analysis via debate}.
\newblock In \emph{Findings of the Association for Computational Linguistics: EMNLP 2023}, page 7572–7590. Association for Computational Linguistics.

\bibitem[{Xue and Salim(2023)}]{xue2023promptcast}
Huan Xue and Flora~D Salim. 2023.
\newblock Promptcast: A new prompt-based learning paradigm for time series forecasting.
\newblock \emph{IEEE Transactions on Knowledge and Data Engineering}.

\bibitem[{Zeng et~al.(2023)Zeng, Chen, Zhang, and Xu}]{zeng2023transformers}
Ailing Zeng, Muxi Chen, Lei Zhang, and Qiang Xu. 2023.
\newblock Are transformers effective for time series forecasting?
\newblock In \emph{Proceedings of the AAAI Conference on Artificial Intelligence}, volume~37, pages 11121--11128.

\bibitem[{Zhang et~al.(2024{\natexlab{a}})Zhang, Du, Shan, Zhou, Du, Tenenbaum, Shu, and Gan}]{zhang2024buildingcooperativeembodiedagents}
Hongxin Zhang, Weihua Du, Jiaming Shan, Qinhong Zhou, Yilun Du, Joshua~B. Tenenbaum, Tianmin Shu, and Chuang Gan. 2024{\natexlab{a}}.
\newblock \href {https://arxiv.org/abs/2307.02485} {Building cooperative embodied agents modularly with large language models}.
\newblock \emph{Preprint}, arXiv:2307.02485.

\bibitem[{Zhang et~al.(2024{\natexlab{b}})Zhang, Xu, Zhang, Liu, Hooi, and Deng}]{Zhang2024@explore}
Jintian Zhang, Xin Xu, Ningyu Zhang, RUibo Liu, Bryan Hooi, and Shumin Deng. 2024{\natexlab{b}}.
\newblock Exploring collaboration mechanisms for llm agents: A social psychology view.
\newblock In \emph{In Proceedings of the 62th Annual Meeting of the Association for Computational Linguistics}, volume~1, pages 14544--14607.

\bibitem[{Zhao et~al.(2024)Zhao, Wang, Zhang, Jin, Zhu, Chen, and Xie}]{zhao2024competeaiunderstandingcompetitiondynamics}
Qinlin Zhao, Jindong Wang, Yixuan Zhang, Yiqiao Jin, Kaijie Zhu, Hao Chen, and Xing Xie. 2024.
\newblock \href {https://openreview.net/forum?id=wGtzp4ZT1n} {Compete{AI}: Understanding the competition dynamics of large language model-based agents}.

\bibitem[{Zheng et~al.(2023)Zheng, Zhang, Nguyen, Rampal, Alawadhi, Rong, Head-Gordon, Borgs, Chayes, and Yaghi}]{Zheng2023}
Zhiling Zheng, Oufan Zhang, Ha~L. Nguyen, Nakul Rampal, Ali~H. Alawadhi, Zichao Rong, Teresa Head-Gordon, Christian Borgs, Jennifer~T. Chayes, and Omar~M. Yaghi. 2023.
\newblock Chatgpt research group for optimizing the crystallinity of mofs and cofs.
\newblock \emph{ACS Central Science}, 9(11):2161--2170.

\bibitem[{Zhou et~al.(2021)Zhou, Zhang, Peng, Zhang, Li, Xiong, and Zhang}]{zhou2021informer}
Haoyi Zhou, Shanghang Zhang, Jieqi Peng, Shuai Zhang, Jianxin Li, Hui Xiong, and Wancai Zhang. 2021.
\newblock Informer: Beyond efficient transformer for long sequence time-series forecasting.
\newblock In \emph{Proceedings of the AAAI Conference on Artificial Intelligence}, volume~35, pages 11106--11115.

\bibitem[{Zhou et~al.(2022{\natexlab{a}})Zhou, Ma, Wen, Sun, Yao, Yin, and Jin}]{zhou2022film}
Tian Zhou, Ziqing Ma, Qingsong Wen, Liang Sun, Tao Yao, Wotao Yin, and Rong Jin. 2022{\natexlab{a}}.
\newblock Film: Frequency improved legendre memory model for long-term time series forecasting.
\newblock \emph{Advances in Neural Information Processing Systems}, 35:12677--12690.

\bibitem[{Zhou et~al.(2022{\natexlab{b}})Zhou, Ma, Wen, Wang, Sun, and Jin}]{zhou2022fedformer}
Tian Zhou, Ziqing Ma, Qingsong Wen, Xue Wang, Liang Sun, and Rong Jin. 2022{\natexlab{b}}.
\newblock Fedformer: Frequency enhanced decomposed transformer for long-term series forecasting.
\newblock In \emph{International Conference on Machine Learning}, pages 27268--27286. PMLR.

\bibitem[{Zhou et~al.(2024)Zhou, Niu, Sun, Jin et~al.}]{zhou2024one}
Tian Zhou, Peisong Niu, Liang Sun, Rong Jin, et~al. 2024.
\newblock One fits all: Power general time series analysis by pretrained lm.
\newblock In \emph{Advances in Neural Information Processing Systems}, volume~36.

\bibitem[{Zhou et~al.(2023)Zhou, Chu, Ruan, Jin, Huang, and Li}]{zhou2023ptse}
Yunyi Zhou, Zhixuan Chu, Yijia Ruan, Ge~Jin, Yuchen Huang, and Sheng Li. 2023.
\newblock Ptse: A multi-model ensemble method for probabilistic time series forecasting.
\newblock In \emph{The 32nd International Joint Conference on Artificial Intelligence}.

\end{thebibliography}

\clearpage
\appendix
\section{Experimental Settings}
\label{sec:experimental_settings}

\subsection{Details of Datasets}
\label{dataset}
The details of the dataset and the news filtering examples corresponding to each dataset are shown in Table~\ref{tab:dataset_news}.

\begin{table*}[htbp]
\centering
\small
\begin{tabular}{|>{\centering\arraybackslash}m{2.3cm}|>{\centering\arraybackslash}m{2.9cm}|>{\centering\arraybackslash}m{2.9cm}|>{\centering\arraybackslash}m{2.9cm}|>{\centering\arraybackslash}m{2.9cm}|}

\hline
\textbf{Datasets} & \textbf{Electricity} & \textbf{Exchange} & \textbf{Traffic} & \textbf{Bitcoin} \\
\hline
\textbf{Time Horizon} & 2019.01-2021.12 & 2019.01-2022.12 & 2015.01-2016.12 & 2019.01-2021.06 \\
\hline
\textbf{Variates} & 19 & 7 & 862 & 18 \\
\hline
\textbf{Timestep} & 52,560 & 1,460 & 17,544 & 858 \\
\hline
\textbf{Granularity} & 30 minutes & 1 day & 1 hour & 1 day \\
\hline
\textbf{Input length} & 48 & 7 & 24 & 7 \\
\hline
\textbf{Prediction length} & 48 & 7 & 24 & 7 \\
\hline
\textbf{Prediction Variable} & load consumption & AUD/USD exchange rate & traffic volume & bitcoin price \\
\hline
\textbf{News examples filtered based on the corresponding datasets} & 
South Australia is only days away from a heatwave which will last for almost a week and has left Tour Down Under organisers anxiously watching the weather forecast. & 
The RBA has dramatically revised down its economic forecasts amid the ongoing property market correction, prompting the Australian dollar to plunge again. & 
A funnel cloud was spotted over Waterford in northern California on April 27 as a line of storms brought heavy rain and hail to the area. & 
Personal finance expert Peter Adeney, known as 'Mr. Money Mustache,' has warned against investing in bitcoin, calling it a speculative asset rather than a true investment. \\
\hline
\end{tabular}
\caption{Dataset Information and News Examples}
\label{tab:dataset_news}
\end{table*}

\subsection{Implementation Details}
We try GLM-4-130B, DeepSeek-V2.5 and GPT-4o for $\text{LLM}_L$. The temperature is set at 0.5, top-k is set at 20 and top-p is set at 0.8. These settings can verify the performance of the competition mechanism under different LLMs. $\text{LLM}_S$ uses LLama-2-7B, the parameter assignments of which are the same with \citet{Wang2024}. All experiments were run on a server equipped with 4 NVIDIA A800 GPUs (80GB each).

During fine-tuning, we applied the LoRa method to Llama 2, setting the LoRa rank to 8 or 16 depending on token length, with alpha = 16 and a learning rate of 0.0001\citep{Wang2024}. Numerical formatting retained three significant digits to avoid excessive tokenization. The fine-tuning was conducted on an NVIDIA A800, with each model instance undergoing hundreds to 1000 training iterations, taking up to a day.

For Deep Neural network baselines, the non-numeric data is taken as dummy variables before fed into baselines. To ensure the reliability of the experiments, for each baseline, we followed the official architecture settings, which are reported in their researches, to assign parameters.

\subsection{Example of Textual Input for Fine-tuning LLM}
\label{input}
The construction of input and output refers to the study by \citet{Wang2024}. The specific input is as follows.

\begin{tcolorbox}[colframe=purple!80!black, colback=white, coltitle=white, title=\centering An Example of Input Data]
        \{
        "instruction": "The historical load data is: 4640.1,4476.7,4343.7,4257.5,4082.8,3923.4, ...",
        
        "input": "Based on the historical load data, please predict the load consumption in the next day. The region for prediction is VIC. The start date of historical data was on 2020-4-9 that is Weekday, and it is not a public holiday. The data frequency is 30 minutes per point. Historical data covers 1 day. The date of prediction is on 2020-4-10 that is Weekday, and it is a public holiday: Good Friday. Weather of the start date: the minimum temperature is 284.96; the maximum temperature is 294.13; the humidity is 87.0; the pressure is 1017.0.  Weather of the prediction date: the minimum temperature is 285.24; the maximum temperature is 291.11; the humidity is 87.0; the pressure is 1005.0. On 2020-04-09, in the state of National, the news was: 'The largest financial package in Australian history has passed through parliament after getting the green light in the Senate on Wednesday night.'. Rationality behind it: The financial stimulus package could lead to increased economic activity, potentially boosting industrial and commercial electricity demand in the long term. ...",
        
        "output": "4741.8,4497.8,4360.1,4188, ..."
        \}
\end{tcolorbox}

\subsection{Information Asymmetry (IA)} 
In a discussion, IA embodies information asymmetry from two aspects: First, IA allows an agent to send information to all agents, or choose to send information to selected agents (\textbf{Selective communication}). As introduced in previous study ~\citep{wu2024shallteamupexploring}, agents will spontaneously cooperate in competition. The mode design can assist agents to adopt more flexible strategies to decide competition or collaboration. Second, IA allows an agent to publish incomplete or misleading logic to other opponents (\textbf{Hide or forge logic}). Information asymmetry is an inherent attribute or strategy to prevent opponents from obtaining a player’s key information. Additionally, research has shown that IA can significantly improve the stability and efficiency of the agents' learning process, outperforming independent learning scenarios \citep{tampubolon2021informationasymmetrycompetitivemultiagent}. IA allows agents to independently determine how much of its news filtering logic to share with competitors in the discussion. Options range from full disclosure of their logic, partial sharing of selected elements, to deliberate fabrication of misleading or incorrect logic. The output of IA is described as below:

\vspace{-0.2cm}

\begin{equation}
\begin{aligned}
& \text{PL}^{e} = \{pl^{e}_1, pl^{e}_2, ..., pl^{e}_I\} \\
& pl^{e}_i = \text{LLM}_L(\mathcal{P}_{\text{IA}}, X, logic^{e}_i, eval^{(e)}, target)
\end{aligned}
\end{equation}

\vspace{-0.1cm}

\noindent where $\text{PL}$ is the set of logic, which are published by each agent in round $e$. $pl^{e}_i$ is the logic and its explanation published by agent $i$. $EM^{(e)}$ contains the evaluation results of all the agents. $target$ signifies the subset of agents with whom agent $i$ desires to initiate communication from the entire pool of agents. The prompt template of IA is $\mathcal{P}_{\text{IA}}$, the detailed description of which could be seen in Prompt 2 of Appendix~\ref{prompt}. $logic^{e}_i$ is the logic of agent $i$ in round $e$. $\mathcal{P}_{\text{IA}}$ aggregates all the inputs to form a comprehensive prompt, and $\text{LLM}_L$ is the large language model to execute the prompt.

\subsection{Opponent-Oriented Self-Reflection (OOSR)} 
After IA component, each agent can update its own news selection logic by referencing the logic $PL^{(e)}$ of others. Due to the presence of incomplete and misleading information in the $PL^{(e)}$, the wrong logic propagation error will be magnified. We propose the MSR model to enhance agents’ ability to discriminate against misleading logic.

\textbf{Multi-Stage Reflection (MSR). }MSR contains three stages. In the first stage, following the method proposed by \citet{Wang2024}, each agent updates the news selection logic, which is expressed as below:

\vspace{-0.5cm}

\begin{equation}
\begin{aligned}
L^{(e+1)'}_i = \text{LLM}_{L}(\mathcal{P}_{\text{ref}},X,\text{PL}^{e}_{-i}, L^{(e)}_i, EM^{(e)})
\end{aligned}
\end{equation}

\noindent where $L^{(e+1)'}_i$ is the updated news selection logic of the $i$th agent in round $e+1$. $\mathcal{P}_{ref}$ is the prompt template, the detailed description of which could be seen in Appendix~\ref{prompt}. $\text{PL}^{e}_{-i}$ is the set of all agents' published logic and explanations except agent $i$. The prompt template aggregates all the inputs to form a comprehensive prompt, and $\text{LLM}_L$ is the LLM to execute the prompt.  

In the second stage, we design a $\text{diff}$ function to extract the updated parts from $L^{(e+1)'}_i$ compared with $L^{(e)}_i$. The formula could be seen as below:

\begin{equation} \label{eq6}
\begin{aligned}
\delta^{(e+1)}_{i} & = \{\delta_1, \delta_2, ..., \delta_U\} \\
& = \text{diff} (L^{(e+1)'}_i, L^{(e)}_i)
\end{aligned}
\end{equation}

\noindent where $\delta^{(e+1)}_{i}$ is the set of $U$ updated parts of agent $i$ in round $e$. For the $u$th updated part $\delta_u$ ($u \leq U$) in $\delta^{(e+1)}_{i}$, we use formula ~(\ref{eq5}) to judge whether it is good logic or not.

\vspace{-0.5cm}

\begin{equation} \label{eq5}
\left\{
\begin{aligned}
& ID(\delta_u) = good, if \ \text{IR}(L^{(e+1)'}_i - \delta_u) \leq \text{IR}(L^{(e+1)'}_i) \\
& ID(\delta_u) = bad, if \ \text{IR}(L^{(e+1)'}_i - \delta_u) > \text{IR}(L^{(e+1)'}_i)
\end{aligned}
\right.
\end{equation}

\noindent where $\text{IR}$ is the function to adopt the fine-tuned $\text{LLM}^{(e)_S}$ to calculate the MAPE score of agent $i$'s performance based on a specific logic. The formula indicates that removing a "good" $\delta_u$ to the logic $L^{(e+1)'}_i$ can decrease the performance, while removing the "bad" one can improve the performance. The significance of designing this formula lies in our use of quantitative indicators to assist LLMs in making judgments about misleading logic (bad one), thereby enhancing the controllability of the reflective process.

In the third stage, We retain all updated parts marked as good in $\text{IR}(L^{(e+1)'}_i)$, and re-evaluate those marked as bad in conjunction with temporal trends to finally determine whether to keep them. Reflection in this stage ensures that an excessive number of updated parts is not discarded. Assume the final removed parts are $\delta^{(e+1)}_{i,bad}$, and the final logic $L^{(e+1)}_{i}$ for the next round of competition is expressed as:

\begin{equation}
\begin{aligned}
L^{(e+1)}_{i} = L^{(e+1)'}_{i} - \delta^{(e+1)}_{i,bad}
\end{aligned}
\end{equation}

\noindent where the minus sign indicates removing $\delta^{(e+1)}_{i,bad}$ from $L^{(e+1)'}_{i}$.

\subsection{Definition of CLD and CPD}
In each round of competition, each agent is required to explain the authenticity of the logic they publish (The detailed prompt design can be seen in Prompt 11 of Appendix H). For an agent, if the logic it publishes to another agent is authentic, then we define its communication in this instance as a collaborative communication. Assume after $E$ rounds of competitions, that the total number of agent $i$'s published logic is $N_{all}$, the number of collaborative communications is $N_{c}$, then we define the collaborative degree of agent $i$ is $\text{CLD} = N_{c}/N_{all}$, and $1 - \text{CLD}$ is the competitive degree (CPD).

\section{Tests of Other LLMs}
\label{sec:LLMs}

\subsection{Tests of Other Small-Scale LLMs Models}
The experimental results on other small-scale LLMs models for fine-tuning are shown Table~\ref{tab:LLMs}. Mistral v0.1\citep{jiang2023mistral}, a 7B model, produced similar results as Llama 2 (7B)\citep{touvron2023llama}. The Gemma 2B model\citep{team2023gemini} had slightly worse results, which may be due to its limited number of parameters. It is necessary to adjust the training for small models to achieve better results. Nonetheless, the results demonstrate the potential of language models to achieve good performance in our proposed methods.

\begin{table}[htbp]
    \centering
    \small
    \setlength{\tabcolsep}{4pt} 
    \renewcommand{\arraystretch}{1.2} 
    \begin{tabular}{lcccc}
        \hline
        \multicolumn{5}{c}{\textbf{Electricity}} \\  
        \hline
        & \textbf{RMSE} & \textbf{MSE}$_{\times10^{-3}}$ & \textbf{MAE} & \textbf{MAPE} \\
        \hline
        Llama 2 (7B)       & 365.52 & 133.60  & 229.19 & 6.71\% \\
        Qwen 2 (7B)        & 371.42 & 137.95  & 278.18 & 7.39\% \\
        Mistral v0.1 (7B)  & 369.71 & 136.69  & 248.44 & 7.21\% \\
        Gemma 2 (2B)       & 370.08 & 136.96 & 236.72 & 6.83\% \\
        \hline
    \end{tabular}
    \caption{Performance comparison on other LLMs}
    \label{tab:LLMs}
\end{table}

\subsection{Tests of Other Large-Scale LLMs models}
The experimental results on other large-scale LLMs models for prompt based reasoning are show in Table~\ref{tab:api}. Three LLMs, GLM-4-130B, DeepSeek-V2.5 and GPT-4o are chosen for comparisons. They play the role as news logic generation, news selection, competition awareness and reflection in the framework. Llama 2 (7B) is taken as $\text{LLM}_S$ to provide prediction assistance for $\text{LLM}_L$. The performance of the three LLMs is relatively close, with GPT-4o showing better results. This indicates that the method proposed in this study can be effectively applied to different large language models.   

\begin{table}[htbp]
    \centering
    \small
    \setlength{\tabcolsep}{4pt} 
    \renewcommand{\arraystretch}{1.2} 
    \begin{tabular}{lcccc}
        \hline
        \multicolumn{5}{c}{\textbf{Electricity}} \\  
        \hline
        & \textbf{RMSE} & \textbf{MSE}$_{\times10^{-3}}$ & \textbf{MAE} & \textbf{MAPE} \\
        \hline
        Deepseek v2.5       & 365.52 & 133.60  & 229.19 & 6.71\% \\
        GLM 4 (130B)       & 378.61 & 143.35  & 249.13 & 7.14\% \\
        GPT 4o   & 363.77 & 132.33  & 218.45 & 6.54\% \\
        \hline
    \end{tabular}
    \caption{Performance comparison on other LLMs}
    \label{tab:api}
\end{table}

\section{Parameter Sensitivity Analysis}
\label{Parameter Sensitivity Analysis}

\subsection{Impact of Retention Ratio on Model Performance}
\label{Impact of Retention Ratio}
\begin{figure}[t]
  \includegraphics[width=\columnwidth]{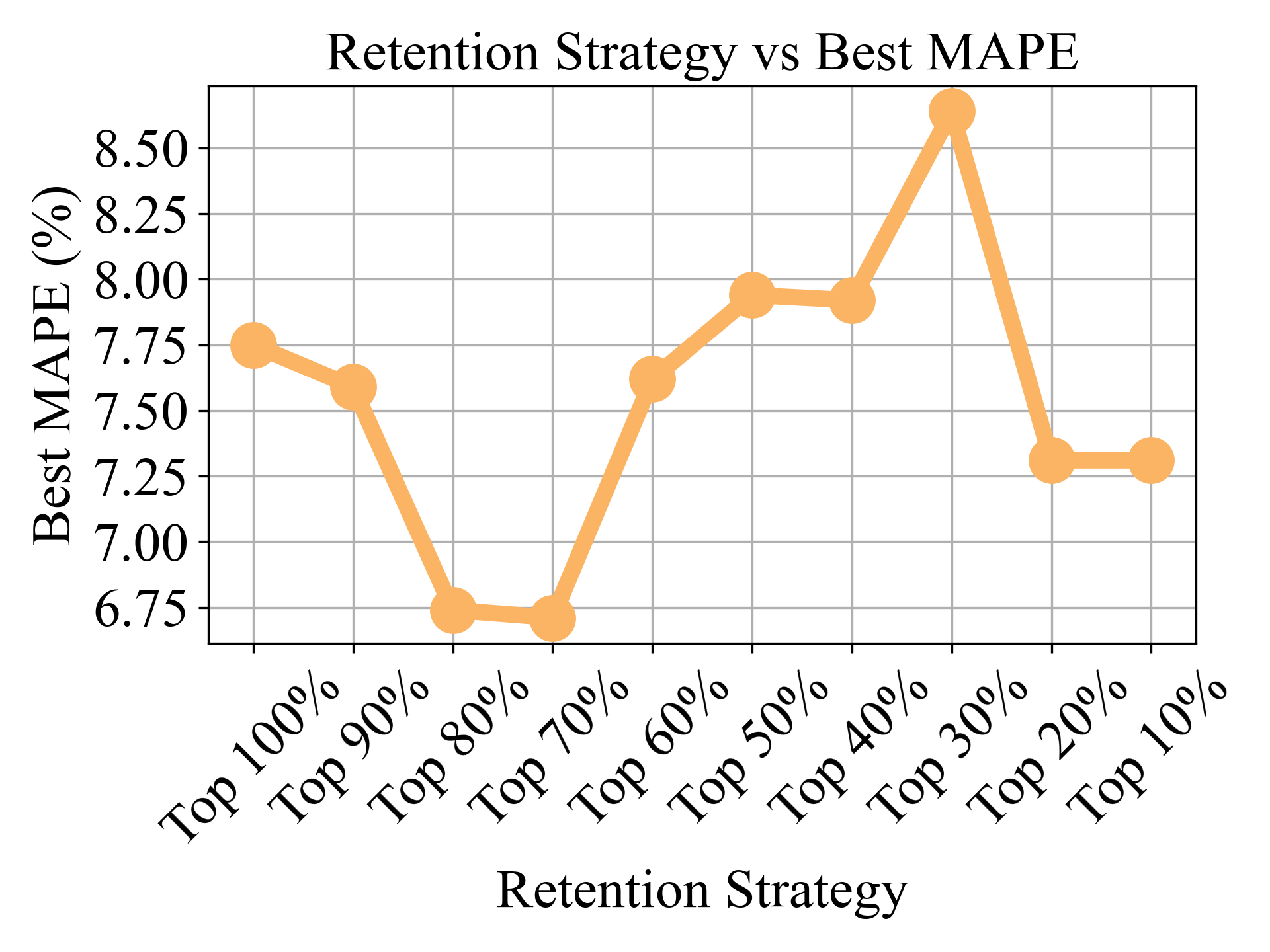} 
  \caption{
The impact of retention rates ranging from 10\% to 90\% on prediction accuracy. Prediction accuracy is measured using MAPE, where a smaller MAPE value indicates higher prediction accuracy.BEST MAPE represents the lowest MAPE across all iterations, which corresponds to the MAPE of the iteration with the highest prediction accuracy.
  }
  \label{fig:retention strategies}
\end{figure}

We compare different $\alpha$ in SF component to explore the impact of the retention ratio on the proposed model. We take the value of $\alpha$ from 0 to 1, and the experimental result on Electricity data is shown in Figure~\ref{fig:retention strategies}. It can be seen that when $\alpha$ is equal to 70\%, the model can obtain the best MAPE score. Therefore, in our experiment, we set the value of $\alpha$ to 0.7, indicating that when SF is triggered, 30\% of the agents with the lowest rankings will be eliminated to ensure that high-performing agents can enter the final group decision-making.

\subsection{Impact of Different Number of Initial Agents}
\label{Different Number of Initial Agents}

\begin{figure}[!ht]
  \includegraphics[width=\columnwidth]{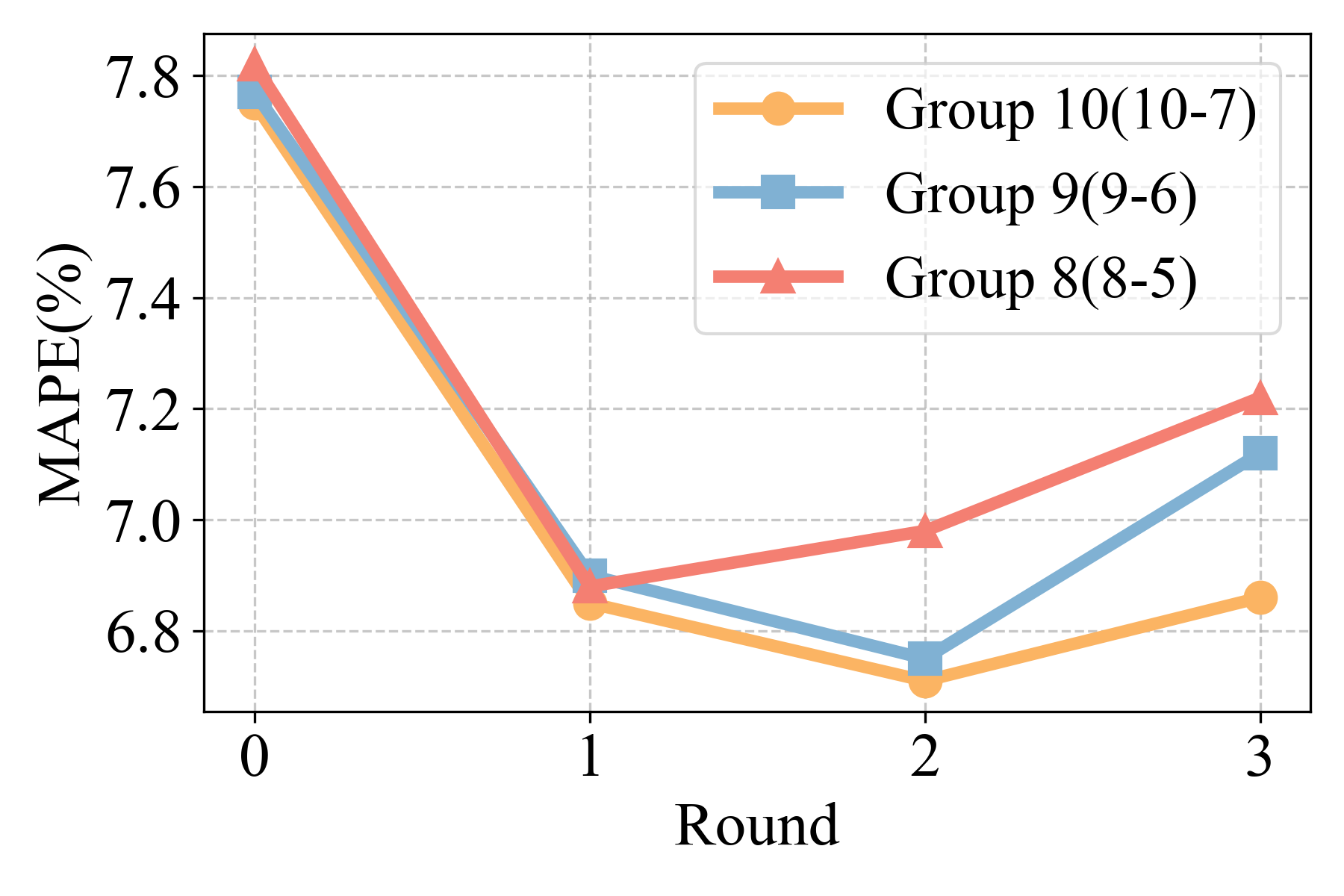} 
  \caption {The figure illustrates the impact of the initial number of competing agents on prediction accuracy. In each epoch, only the \textbf{top 70\% of agents} are retained. We can observe that all models with different initial agent number settings reached their optimal values at epoch 2. Thus, if the initial number of competing agents is 10, the number of agents evolves as follows over three iterations: 10 → 7 → 4.}
  \label{init_num}
\end{figure}

We compare different number of initial agents to explore the impact of this parameter on the proposed model. Due to the limitation of computational hardware (4 A800 GPU cards), we set the maximum number of agents to 10. We take the number from 8 to 10, and the experimental result on Electricity data is shown in Figure~\ref{init_num}. It can be observed that all models with different initial agent population settings reached their optimal values at epoch 2. The model with an initial number of 10 agents performs the best. The model retained the top 4 performing agents for group decision-making at epoch 2 (counting from epoch 0). This demonstrates the effectiveness of the model’s competitive and elimination mechanisms. Subsequently, as the number of agents further decreased, the model’s performance weakened, which to some extent indicates that the model experienced over-competition. Based on the aforementioned experimental observations, we set the initial number of agents to 10, with the entire training process consisting of 3 rounds of epochs (1, 2, 3).

\subsection{Impact of Temperature on Model Performance}
\label{Impact of Temperature}

\begin{figure}[!ht]
  \includegraphics[width=\columnwidth]{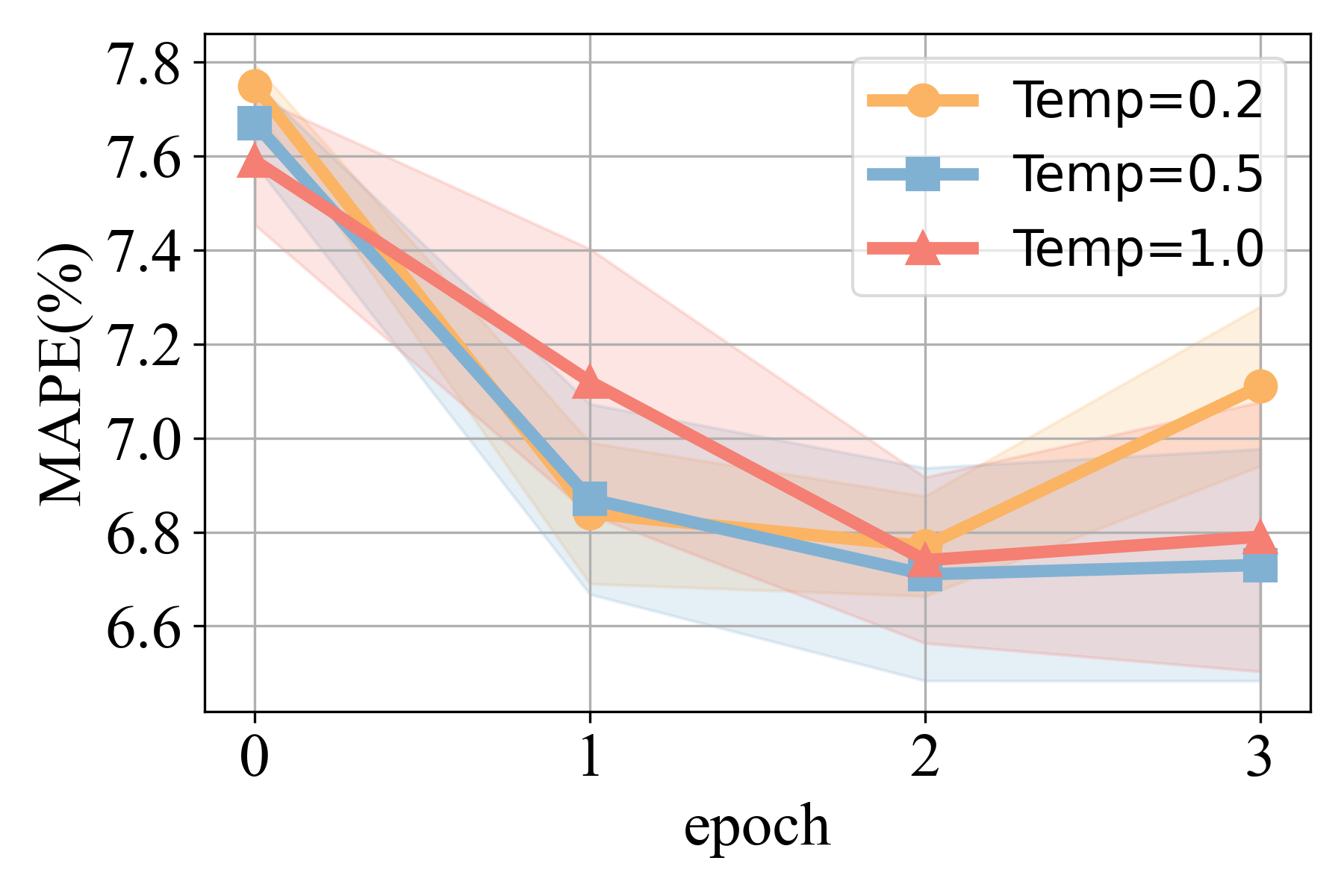} 
  \caption{
The impact of temperature on the model's performance.}
  \label{fig:temperature}
\end{figure}

We set each agent at different temperatures and repeated the experiments. The experimental results are shown in Figure~\ref{fig:temperature}. It can be seen that the increase in temperature may cause the variance of the model's prediction results to grow, but overall, the iterative competitive mechanism still improves the model's accuracy. In another aspect, after multiple rounds of testing, we found that the variance can be well controlled within a small range, indicating that the randomness introduced by temperature does not significantly affect the randomness of the results, and the overall performance of the model is relatively stable. This indicates that the improvement in model accuracy due to the iterative competitive mechanism is not significantly affected by the temperature. We set the temperature at 0.5.

\subsection{Impact of Competitive Intensity Coefficient}
\label{sec:CI}

\begin{figure}[t]
  \includegraphics[width=\columnwidth]{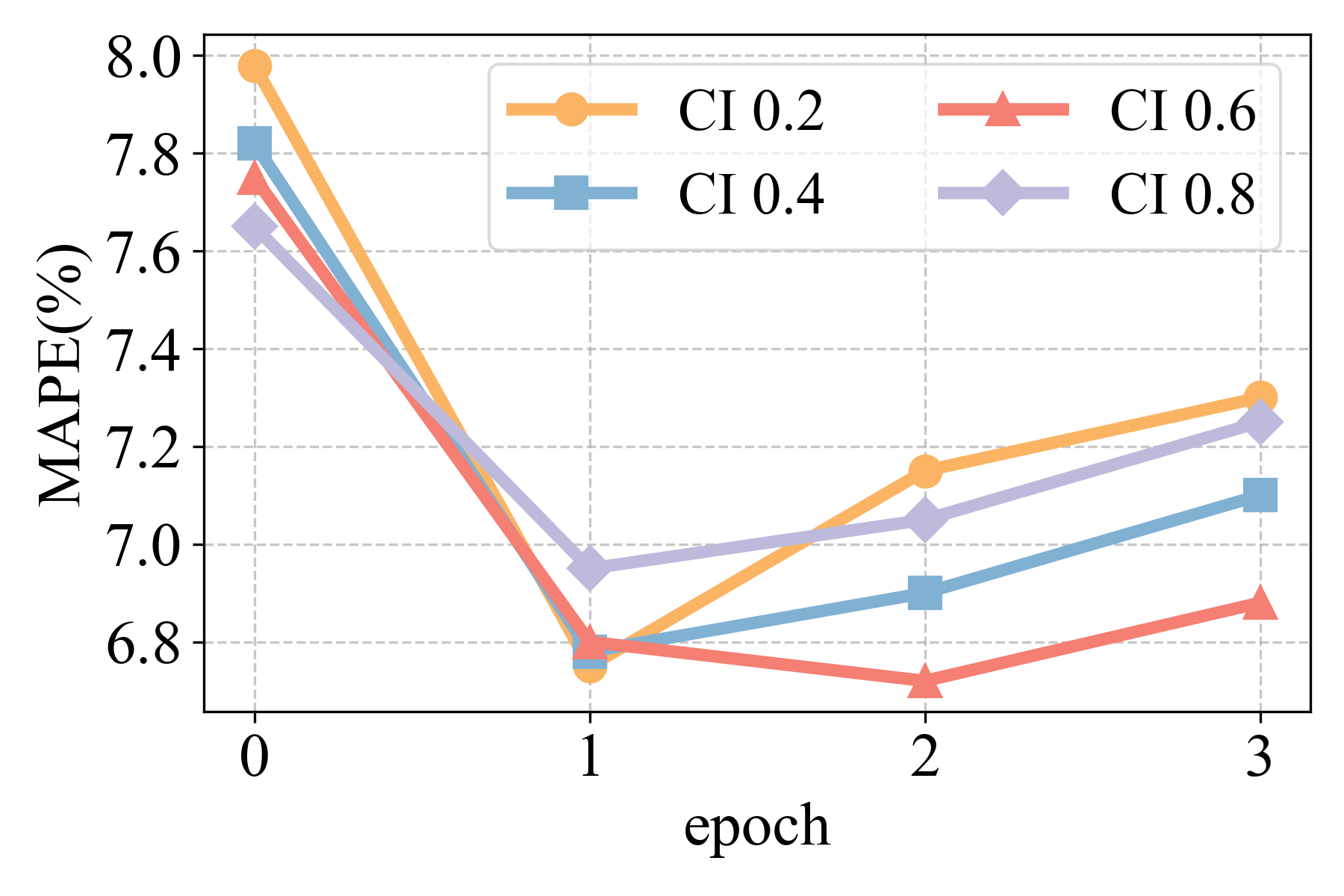} 
  \caption{
The variation in the competition index affects the accuracy of model predictions. Round represents each iteration.}
  \label{fig:CI}
\end{figure}

In the sensitivity analysis of the Competitive Intensity Coefficient (CI), we categorize all agents into high-competitiveness and low-competitiveness groups, where CI represents the proportion of high-competitiveness agents in the total population. Compared to low-competitiveness agents, high-competitiveness agents are more inclined to conceal part of their true filtering logic and may fabricate misleading logic to interfere with their opponents’ judgments. We set the total number of agents to 10 and conducted experiments under different CI values: 0.2 (20\% high-competitiveness agents), 0.4 (40\%), 0.6 (60\%), and 0.8 (80\%), with the results shown in Figure~\ref{fig:CI}.

 The results indicate that when \( CI = 0.6 \), the Mean Absolute Percentage Error (MAPE) is at its lowest in the second and third iterations, while in the first iteration, the MAPE value is also close to the optimal level. When CI deviates from 0.6, either decreasing or increasing, the MAPE values in all iterations tend to rise to varying degrees. Furthermore, as the number of iterations increases, the upward trend in MAPE becomes more pronounced. This may be because when the number of high-competitiveness agents is too low, agents tend to fully disclose their true filtering logic to their opponents, leading to convergence in filtering logic and a lack of diversity in the optimization process. Conversely, when the proportion of high-competitiveness agents is too high, agents are more likely to conceal their true logic and fabricate misleading logic, which can not only reduce the optimization effectiveness of their opponents but also mislead them into selecting irrelevant news, ultimately degrading the overall filtering quality. In general, our framework can consistently work under different competition intense.

 \section{Models for comparison}
\begin{itemize}
    \item \textbf{Autoformer} \citep{wu2021autoformer}: Autoformer addresses long-term time series forecasting by introducing a novel decomposition architecture integrated into the Transformer framework.  It replaces traditional self-attention with an Auto-Correlation mechanism based on series periodicity, enhancing both efficiency and accuracy for long-term predictions. This model innovatively incorporates decomposition as an inner block, enabling progressive decomposition capabilities.

    \item \textbf{Informer} \citep{zhou2021informer}: Informer is designed to be an efficient Transformer for long sequence time series forecasting. It tackles the limitations of standard Transformers by introducing ProbSparse self-attention (reducing complexity), self-attention distilling (handling long inputs), and a generative-style decoder (improving inference speed). These innovations make Transformers more practical for long sequence forecasting.

    \item \textbf{DLinear} \citep{zeng2023transformers}: DLinear challenges the prevalent use of Transformers for time series forecasting. It introduces simple one-layer linear models (LTSF-Linear) and demonstrates that these surprisingly outperform complex Transformer-based models on various datasets. The work questions the effectiveness of self-attention in capturing temporal relations in time series data.

    \item \textbf{iTransformer} \citep{liu2023itransformer}: iTransformer proposes an inverted Transformer architecture for time series forecasting.  It applies attention and feed-forward networks on inverted dimensions, allowing the model to capture multivariate correlations by attending to variate tokens (series) instead of temporal tokens (time points). This approach aims to improve performance, generalization, and utilization of long lookback windows.

    \item \textbf{FiLM} (Frequency Improved Legendre Memory Model) \citep{zhou2022film}: FiLM focuses on enhancing the preservation of historical information in neural networks for long-term forecasting. It employs Legendre polynomial projections to approximate historical data and Fourier projection to mitigate noise.  FiLM is designed to improve the accuracy of existing models and can be used as a plugin module.

    \item \textbf{Pyraformer} \citep{liu2021pyraformer}: Pyraformer introduces Pyramidal Attention Module (PAM) to explore multi-resolution representations of time series. It uses an inter-scale tree structure and intra-scale neighboring connections to capture temporal dependencies efficiently.  Pyraformer achieves linear time and space complexity, making it suitable for long-range time series modeling and forecasting.

    \item \textbf{PatchTST} \citep{nie2022transformers}: PatchTST proposes segmenting time series into patches as input tokens for Transformers. It employs channel-independence, where each channel is processed independently with shared weights.  This patching strategy improves efficiency, retains local semantic information, allows attending to longer history, and enhances long-term forecasting accuracy.

    \item \textbf{FEDformer} \citep{zhou2022fedformer}: FEDformer integrates seasonal-trend decomposition with Transformers for long-term time series forecasting. It uses decomposition to capture the global profile of time series and Transformers to model detailed structures.  Furthermore, it incorporates frequency enhancement based on Fourier transform to improve Transformer performance and efficiency, achieving linear complexity.

    \item \textbf{GPT4TS} \citep{zhou2024one}: GPT4TS explores the use of pre-trained models from NLP and CV for general time series analysis. It introduces the Frozen Pretrained Transformer (FPT), which leverages pre-trained language or image models by freezing their Transformer layers and fine-tuning them for time series tasks.  This work demonstrates the potential of transfer learning and general-purpose models in time series analysis.
\end{itemize}

The visualized results(see Figure~\ref{baseline}) reveal that our model ("Ours") exhibits satisfactory performance in tracking the actual time series data, demonstrating relatively low error margins. While all models display a degree of lag and peak underestimation, "Ours" achieves superior overall prediction accuracy compared to the baseline models. This indicates a potential strength of "Ours" in capturing the dynamic properties of the time series.

\begin{figure*}[t]
  \includegraphics[width=\linewidth]{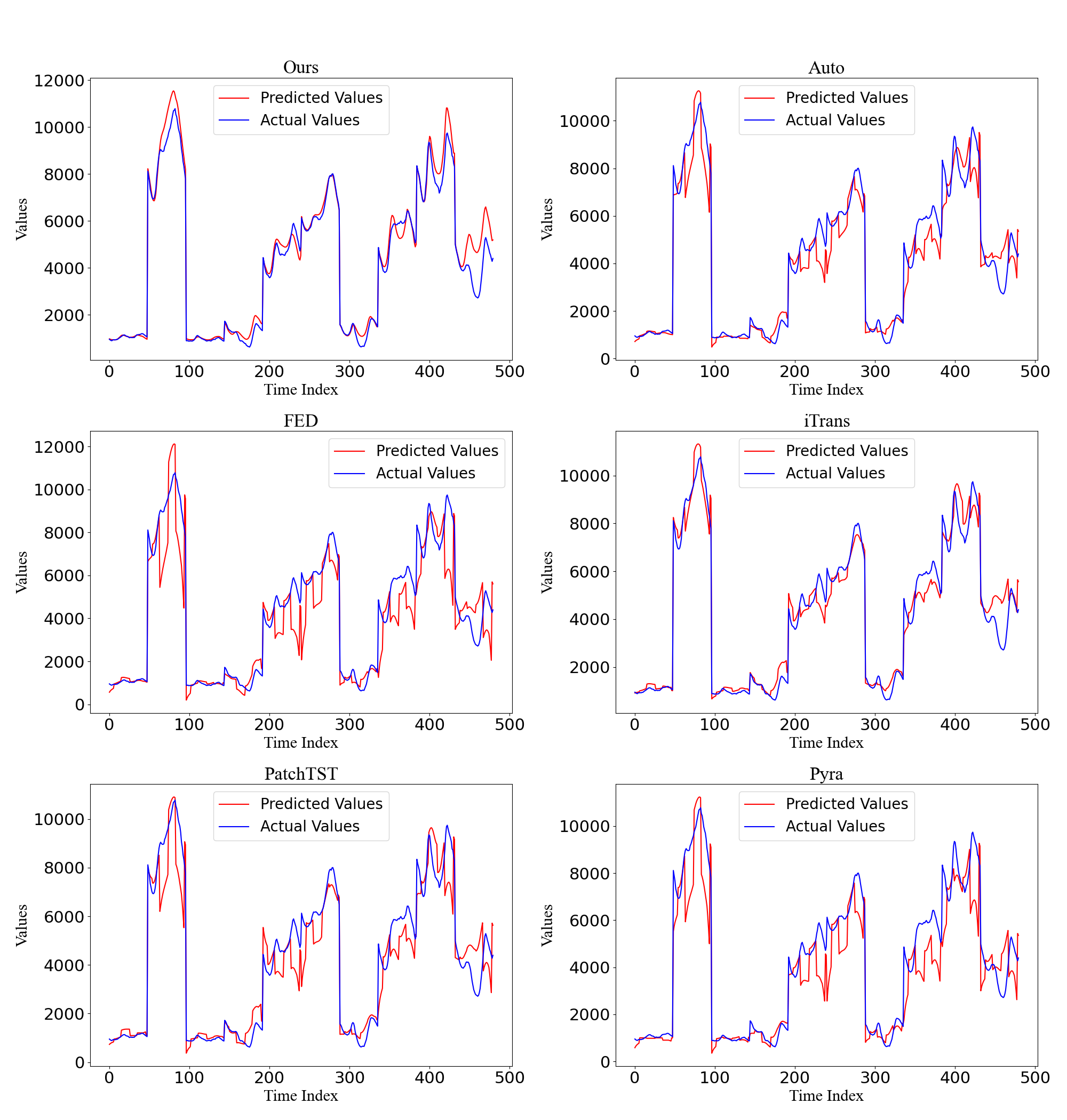} 
  \caption{
    The visualized results of our work and part of the baseline models.
  }
  \label{fig:baseline}
\end{figure*}

\section{Varying Prompt Settings}
\label{sec:prompt}

Our competitive mechanism consists of four components:
\begin{itemize}
    \item Information Asymmetry
    \item Reward and Evaluation Mechanisms
    \item Self-Reflection and Optimization of Agents
    \item Survival of the Fittest
\end{itemize}

In the actual experiments, only the first three components—Information Asymmetry, Reward and Evaluation Mechanisms, and Self-Reflection and Optimization of Agents—are influenced by the prompt settings. Therefore, we used GPT-4 to paraphrase the prompts for these three sections, rephrasing the original prompts in a different form. The specific details of the paraphrasing are shown in Appendix~\ref{prompt}. 

\begin{figure*}[t]
  \begin{subfigure}[b]{0.32\linewidth}
    \includegraphics[width=\linewidth]{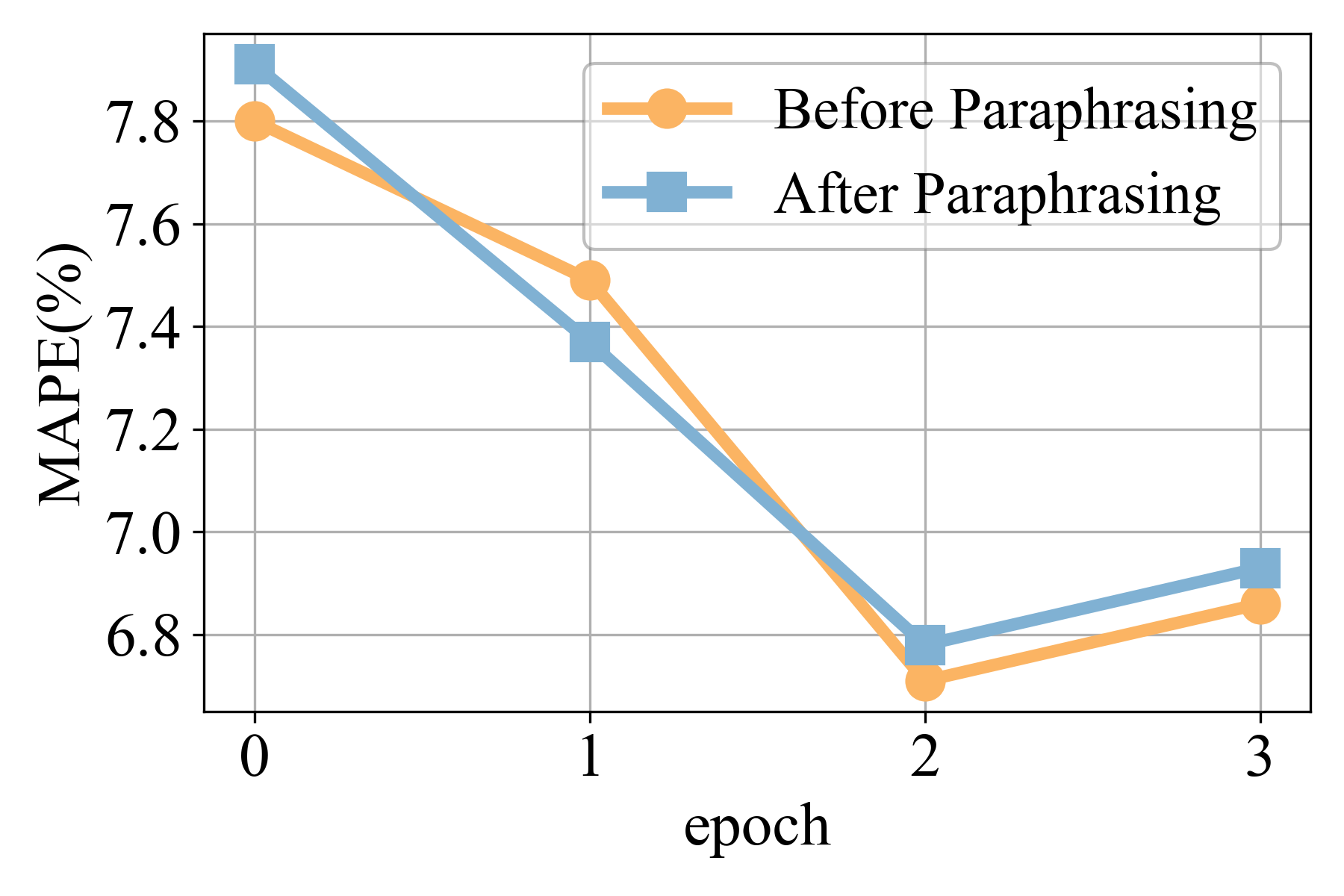}
    \caption{Information Asymmetry} 
  \end{subfigure}
  \hfill
  \begin{subfigure}[b]{0.32\linewidth}
    \includegraphics[width=\linewidth]{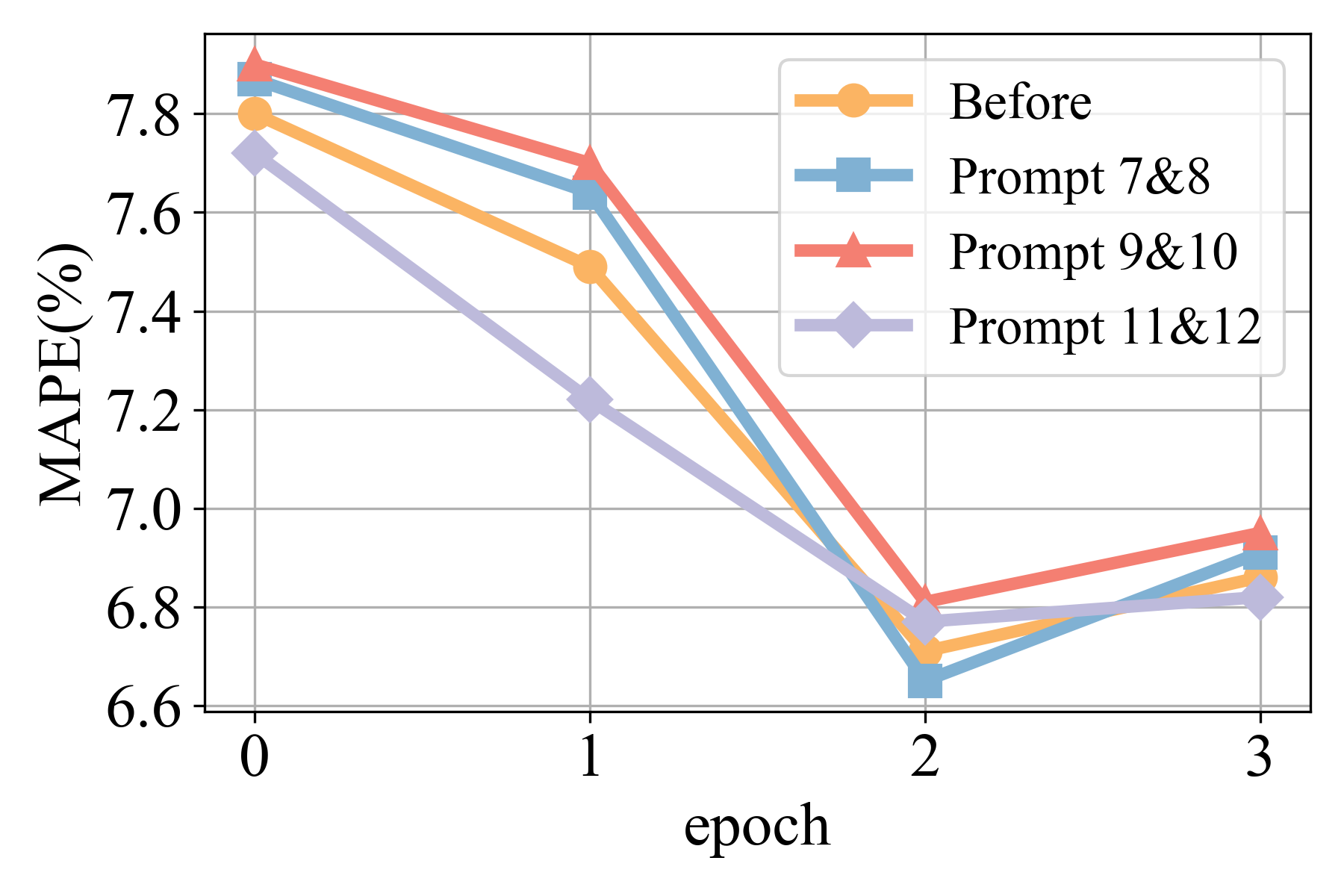}
    \caption{Reward and Evaluation Mechanisms} 
  \end{subfigure}
  \hfill
  \begin{subfigure}[b]{0.32\linewidth}
    \includegraphics[width=\linewidth]{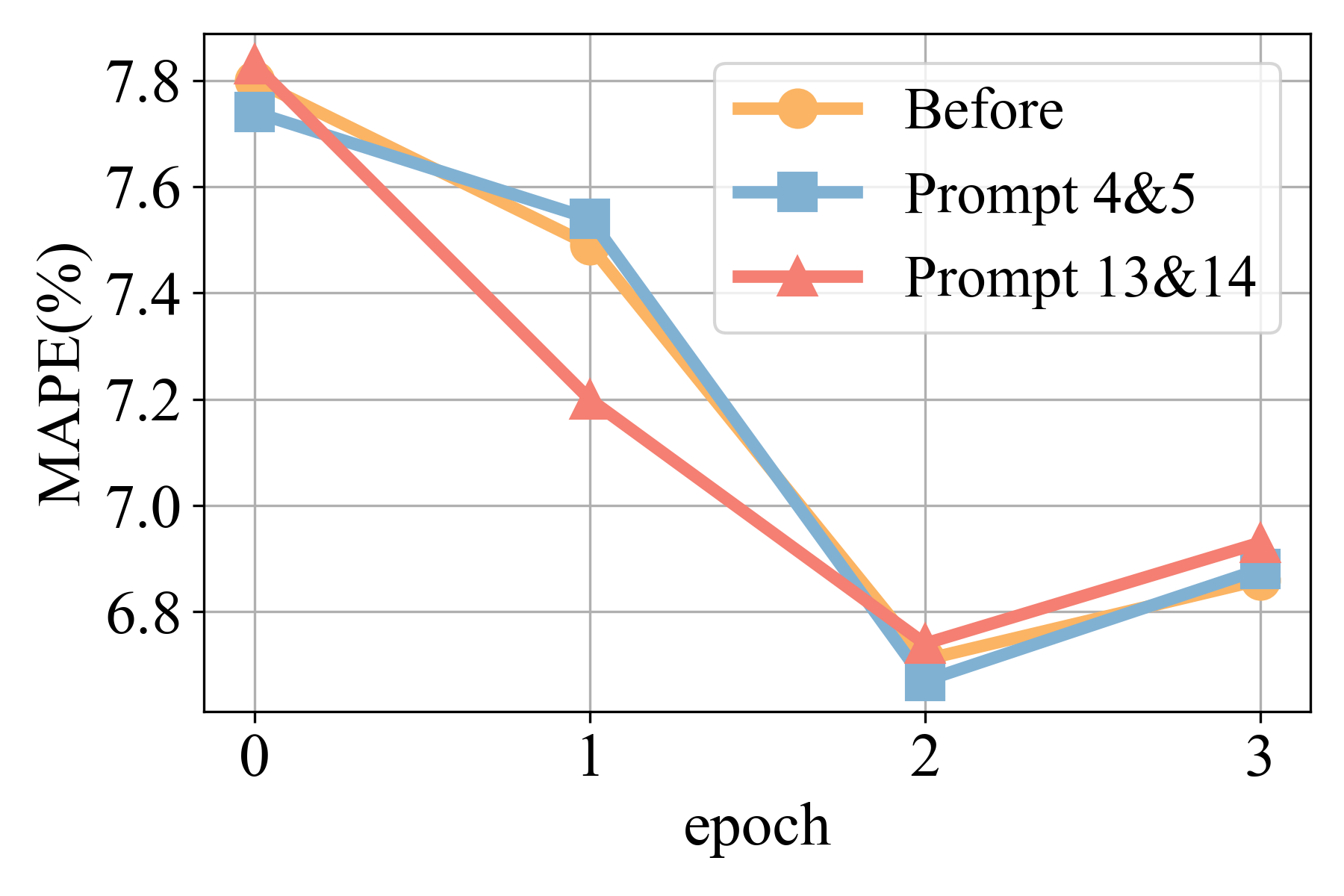}
    \caption{self-Reflection and Optimizations} 
  \end{subfigure}
  \caption{This figure illustrates the effect of prompt rewriting on model prediction accuracy across various competitive modules. Before represents the model's performance prior to any prompt rewriting. Prompt x\&y shows the model's performance after paraphrasing the prompts from Prompt x to Prompt y. }
  \label{prompt_settings}
\end{figure*}

We compared the model prediction accuracy before and after the paraphrasing of prompts for the sections \text{Information Asymmetry} (see Prompt 2 and Prompt 3 in Appendix H), \text{Reward and Evaluation Mechanisms} (see Prompt 7, Prompt 8, Prompt 9, Prompt 10, Prompt 11 and Prompt 12 in Appendix H), and \text{Opponent-orient self-Reflection} (see Prompt 4, Prompt 5, Prompt 13, Prompt 14 in Appendix H). The experimental results are shown in Figure~\ref{prompt_settings}. The impact of prompt paraphrasing on model performance is minimal, further demonstrating the robustness of the model.

\section{Iterative Effects of the Competition Mechanism}
\label{Iterative Effects}
To evaluate the extent of competition's contribution to the ongoing enhancement of news filtering and reflective reasoning, we observe the complete training and optimization process, utilizing epochs as the key observation markers. As seen in Table~\ref{tab:specific_iteration} in Appendix~\ref{Iterative Effects}, the competition mechanism refines news filtering through an iterative process, which is reflected in the progressively improved time series prediction results. To more accurately describe the role of the competition in the training process, we remove the competition mechanism from the entire prediction process and replace it with the discussion framework from \citet{Wang2024@rethinking}, and then re-conducted the experiment. The results are shown in Table 8 in Appendix~\ref{Iterative Effects}. By comparison, it can be observed that at each epoch, the competition mechanism significantly outperforms the discussion mechanism on all evaluation metrics. 

\begin{table*}
\centering
\scriptsize
\begin{tabular}{@{}l|cccccccccccc|cccc@{}}
\cmidrule(r){1-17}
\multicolumn{1}{l|}{\multirow{2}{*}{}}      & \multicolumn{12}{c|}{\textbf{Electricity}}                                                                                                               & \multicolumn{4}{c}{\textbf{Exchange}}                                                                                                              \\ \cmidrule(lr){2-17}
\multicolumn{1}{l|}{}                       & \multicolumn{3}{c|}{RMSE}           & \multicolumn{3}{c|}{MSE (\( \times 10^{-3} \))} & \multicolumn{3}{c|}{MAE}           &  \multicolumn{3}{c|}{MAPE (\( \times 10^{2} \))}          & \multicolumn{1}{c|}{RMSE (\( \times 10^{2} \))}          & \multicolumn{1}{c|}{MSE (\( \times 10^{3} \))} & \multicolumn{1}{c|}{MAE (\( \times 10^{2} \))}          & MAPE (\( \times 10^{2} \))          \\ \cmidrule(r){1-17}
Initial State           & \multicolumn{3}{c|}{469.6}          & \multicolumn{3}{c|}{220.82}               & \multicolumn{3}{c|}{258.72}          & \multicolumn{3}{c|}{7.67}               & \multicolumn{1}{c|}{7.07}        & \multicolumn{1}{c|}{5.00}               & \multicolumn{1}{c|}{2.39}       & 3.25               \\
First Epoch           & \multicolumn{3}{c|}{439.22}          & \multicolumn{3}{c|}{192.91}               & \multicolumn{3}{c|}{250.58}          & \multicolumn{3}{c|}{7.49}               & \multicolumn{1}{c|}{5.33}        & \multicolumn{1}{c|}{2.85}               & \multicolumn{1}{c|}{1.9}       & 2.58               \\
Second Epoch & \multicolumn{3}{c|}{364.52}          & \multicolumn{3}{c|}{132.93}               & \multicolumn{3}{c|}{229.19}          & \multicolumn{3}{c|}{6.71}                & \multicolumn{1}{c|}{3.10}        & \multicolumn{1}{c|}{0.96}               & \multicolumn{1}{c|}{1.25}       & 1.72               \\
Third Epoch & \multicolumn{3}{c|}{439.8}          & \multicolumn{3}{c|}{193.42}               & \multicolumn{3}{c|}{245.65}          & \multicolumn{3}{c|}{6.86}              & \multicolumn{1}{c|}{0.61}        & \multicolumn{1}{c|}{0.037}               & \multicolumn{1}{c|}{0.44}       & 0.63               \\
\cmidrule(r){1-17}
\multirow{2}{*}{}                           & \multicolumn{12}{c|}{\textbf{Traffic}}                                                                                                                   & \multicolumn{4}{c}{\textbf{Bitcoin}}                                                                                                               \\ \cmidrule(lr){2-17}
                                            \multicolumn{1}{l|}{}                       & \multicolumn{4}{c|}{RMSE (\( \times 10^{2} \))}           & \multicolumn{4}{c|}{MSE (\( \times 10^{3} \))} & \multicolumn{4}{c|}{MAE (\( \times 10^{2} \))}           & \multicolumn{1}{c|}{RMSE}          & \multicolumn{1}{c|}{MSE (\( \times 10^{-5} \))} & \multicolumn{1}{c|}{MAE}          & MAPE (\( \times 10^{2} \))          \\ \cmidrule(r){1-17}
Initial State           & \multicolumn{4}{c|}{3.26}          & \multicolumn{4}{c|}{1.06}               & \multicolumn{4}{c|}{1.50}          & \multicolumn{1}{c|}{416.05}        & \multicolumn{1}{c|}{1.73}               & \multicolumn{1}{c|}{278.24}       & 3.14               \\
First Epoch            & \multicolumn{4}{c|}{3.18}          & \multicolumn{4}{c|}{1.01}               & \multicolumn{4}{c|}{1.51}          & \multicolumn{1}{c|}{393.58}        & \multicolumn{1}{c|}{5.5}               & \multicolumn{1}{c|}{264.68}       & 3.06               \\
Second Epoch & \multicolumn{4}{c|}{3.17}          & \multicolumn{4}{c|}{1.00}               & \multicolumn{4}{c|}{1.55}          & \multicolumn{1}{c|}{380.99}        & \multicolumn{1}{c|}{1.45}               & \multicolumn{1}{c|}{255.96}       & 2.92               \\
Third Epoch & \multicolumn{4}{c|}{3.21}          & \multicolumn{4}{c|}{1.03}               & \multicolumn{4}{c|}{1.56}          & \multicolumn{1}{c|}{371.39}        & \multicolumn{1}{c|}{1.38}               & \multicolumn{1}{c|}{248.69}       & 2.83               \\
\cmidrule(r){1-17}
\end{tabular}
  \caption{Comparison of time-series prediction performance across different scenarios with the complete competitive framework. Bold font indicates the categories.}
  \label{tab:specific_iteration}
\end{table*}

\begin{table*}
\centering
\scriptsize
\begin{tabular}{@{}l|cccccccccccc|cccc@{}}
\cmidrule(r){1-17}
\multicolumn{1}{l|}{\multirow{2}{*}{}}      & \multicolumn{12}{c|}{\textbf{Electricity}}                                                                                                               & \multicolumn{4}{c}{\textbf{Exchange}}                                                                                                              \\ \cmidrule(lr){2-17}
\multicolumn{1}{l|}{}                       & \multicolumn{3}{c|}{RMSE}           & \multicolumn{3}{c|}{MSE (\( \times 10^{-3} \))} & \multicolumn{3}{c|}{MAE}           &  \multicolumn{3}{c|}{MAPE (\( \times 10^{2} \))}          & \multicolumn{1}{c|}{RMSE (\( \times 10^{2} \))}          & \multicolumn{1}{c|}{MSE (\( \times 10^{3} \))} & \multicolumn{1}{c|}{MAE (\( \times 10^{2} \))}          & MAPE (\( \times 10^{2} \))          \\ \cmidrule(r){1-17}
Initial State           & \multicolumn{3}{c|}{480.58}          & \multicolumn{3}{c|}{230.96}               & \multicolumn{3}{c|}{274.40}          & \multicolumn{3}{c|}{8.60}               & \multicolumn{1}{c|}{18.16}        & \multicolumn{1}{c|}{32.99}               & \multicolumn{1}{c|}{8.72}       & 12.28               \\
First Epoch           & \multicolumn{3}{c|}{472.86}          & \multicolumn{3}{c|}{223.60}               & \multicolumn{3}{c|}{268.70}          & \multicolumn{3}{c|}{7.83}               & \multicolumn{1}{c|}{10.613}        & \multicolumn{1}{c|}{11.26}               & \multicolumn{1}{c|}{6.25}       & 8.80               \\
Second Epoch & \multicolumn{3}{c|}{418.98}          & \multicolumn{3}{c|}{175.55}               & \multicolumn{3}{c|}{247.71}          & \multicolumn{3}{c|}{7.62}                & \multicolumn{1}{c|}{8.11}        & \multicolumn{1}{c|}{6.57}               & \multicolumn{1}{c|}{5.29}       & 7.42               \\
Third Epoch & \multicolumn{3}{c|}{417.87}          & \multicolumn{3}{c|}{174.62}               & \multicolumn{3}{c|}{250.71}          & \multicolumn{3}{c|}{7.60}              & \multicolumn{1}{c|}{8.66}        & \multicolumn{1}{c|}{7.50}               & \multicolumn{1}{c|}{5.65}       & 7.87               \\
\cmidrule(r){1-17}
\multirow{2}{*}{}                           & \multicolumn{12}{c|}{\textbf{Traffic}}                                                                                                                   & \multicolumn{4}{c}{\textbf{Bitcoin}}                                                                                                               \\ \cmidrule(lr){2-17}
                                            \multicolumn{1}{l|}{}                       & \multicolumn{4}{c|}{RMSE (\( \times 10^{2} \))}           & \multicolumn{4}{c|}{MSE (\( \times 10^{3} \))} & \multicolumn{4}{c|}{MAE (\( \times 10^{2} \))}           & \multicolumn{1}{c|}{RMSE}          & \multicolumn{1}{c|}{MSE (\( \times 10^{-5} \))} & \multicolumn{1}{c|}{MAE}          & MAPE (\( \times 10^{2} \))          \\ \cmidrule(r){1-17}
Initial State           & \multicolumn{4}{c|}{7.41}          & \multicolumn{4}{c|}{5.48}               & \multicolumn{4}{c|}{3.68}          & \multicolumn{1}{c|}{459.71}        & \multicolumn{1}{c|}{2.11}               & \multicolumn{1}{c|}{291.17}       & 3.39               \\
First Epoch            & \multicolumn{4}{c|}{2.39}          & \multicolumn{4}{c|}{0.57}               & \multicolumn{4}{c|}{1.4529}          & \multicolumn{1}{c|}{428.60}        & \multicolumn{1}{c|}{1.84}               & \multicolumn{1}{c|}{261.44}       & 2.98               \\
Second Epoch & \multicolumn{4}{c|}{2.61}          & \multicolumn{4}{c|}{0.68}               & \multicolumn{4}{c|}{1.42}          & \multicolumn{1}{c|}{537.04}        & \multicolumn{1}{c|}{2.88}               & \multicolumn{1}{c|}{296.24}       & 3.26               \\
Third Epoch & \multicolumn{4}{c|}{2.66}          & \multicolumn{4}{c|}{0.71}               & \multicolumn{4}{c|}{1.50}          & \multicolumn{1}{c|}{485.53}        & \multicolumn{1}{c|}{2.36}               & \multicolumn{1}{c|}{296.49}       & 3.40               \\
\cmidrule(r){1-17}
\end{tabular}
  \caption{Comparison of time-series prediction performance across different scenarios without the complete competitive framework. Bold font indicates the categories.}
  \label{tab:specific_iteration_without}
\end{table*}


\section{The construction of Memory Database}
\label{memory}

In the investment decision-making simulation, each agent is equipped with an independent memory bank, which is built based on the memory module in agentscope\citep{agentscope}. By default, it stores all session records and is used for storing and retrieving historical information to assist in decision-making. This memory bank is established when the agent is initialized. At the beginning of each round of conversation, the agent receives a task prompt from the system, which provides background information on the current scenario, including the agent's profit and loss record prior to the current investment and the total number of likes across all agents. During the social interaction phase, the agent will comment and observe other agents' comments, deciding whether to vote for (like) other agents' comments. If a "like" is given, the liked content is updated and stored in the agent's memory bank in the form of a like memory message; interactions without a like are not stored. Meanwhile, the news filtering logic's memory module operates independently of the base memory bank, conducting long-term storage. It aligns and updates based on all useful information in the current round's temporary memory, with iterative updates being overwritten. However, this memory module does not disappear when the base memory bank is cleared. After each round, the agent's temporary memory bank is cleared to prepare for the next round. By simultaneously building long-term and temporary memories, this design effectively controls the length of the context, ensuring that each agent can store and utilize historical information in a personalized manner. Ultimately, all key data, such as comments, likes, and investment decisions, are persistently stored for subsequent analysis. The design of this memory bank aims to assist the agent in making wiser choices in the complex environment of investment decision-making.

\onecolumn 

\section{Case Study}
\label{baseline}
A case study is proposed in this section: on October 25, 2020, the news selected by the agent without a competitive mechanism and the agent with a competitive mechanism, which might influence electricity load changes in NSW, can be seen as follows. Compared with the agent without a competitive mechanism, the agent with the competitive mechanism can screen out more news that aligns with the analytical logic, which may influence electricity load changes in NSW. This further confirms the improvement of the model in terms of diversity in innovative thinking and the ability to judge misleading information.

\begin{tcolorbox}[colframe=purple!80!black, colback=white, coltitle=white, title=\centering News selection results without competitive mechanism]
(1) On 2020-10-25, in the state of National, the news was: 'Early adoption of renewables has Australian-owned hydro, wind and solar schemes helping power economic recovery and employment.'. 

\textbf{Rationality behind it}: The shift toward renewable energy projects and early adoption signifies a long-term change in the energy mix in Australia, potentially reducing reliance on traditional electricity grids while fostering growth in green energy sectors.

\vspace{10pt}

(2) On 2020-10-25, in the state of NSW, the news was: 'Sydney is set to be bombarded with heavy rain on Sunday evening, and experts are undecided which side the wet conditions will favour.'. 

\textbf{Rationality behind it}: The anticipated heavy rainfall can affect today's load consumption by impacting outdoor events and activities leading to an increase in indoor electricity usage as people stay indoors.
\end{tcolorbox}

\begin{tcolorbox}[colframe=purple!80!black, colback=white, coltitle=white, title=\centering News selection results with competitive mechanism]
(1) On 2020-10-25, in the state of National, the news was:
"Early adoption of renewables has Australian-owned hydro, wind, and solar schemes helping power economic recovery and employment."

\textbf{Rationality behind it}: The shift toward renewable energy projects and early adoption signifies a long-term change in the energy mix in Australia, potentially reducing reliance on traditional electricity grids while fostering growth in green energy sectors.

\vspace{10pt}

(2) On 2020-10-25, in the state of National, the news was:
"Australia’s leading real estate identity John McGrath has pinpointed the suburbs he believes will boom in a post-coronavirus world."

\textbf{Rationality behind it}: The prediction of suburban growth in a post-coronavirus world suggests long-term increases in residential and commercial electricity demand as new developments and businesses expand.

\vspace{10pt}

(3) On 2020-10-25, in the state of National, the news was:
"The government has made a call on a confronting Netflix film that some think is a challenging work of art, while others allege it’s a 'pedo film'."

\textbf{Rationality behind it}: This news indirectly highlights the growing importance of digital entertainment, which could lead to increased electricity demand from data centers and streaming services in the long term.

\vspace{10pt}

(4) On 2020-10-25, in the state of NSW, the news was:
"Sydney is set to be bombarded with heavy rain on Sunday evening, and experts are undecided which side the wet conditions will favour."

\textbf{Rationality behind it}: Heavy rain could lead to increased electricity demand for heating or lighting, depending on the temperature drop, resulting in a short-term impact on load consumption.

\end{tcolorbox}


\section{Full Prompt Design}
\label{prompt}

This section details all prompts designed and implemented within the proposed model. To ensure the robustness and stability of our model's performance, the original prompt (designated as "before paraphrasing") was rewritten to create a paraphrased version (designated as "after paraphrasing").  This approach allows us to evaluate the model's sensitivity to variations in prompt wording and confirm its consistent behavior across different phrasings conveying the same underlying intent.  The specific prompts and their corresponding paraphrased versions are presented below.


\begin{tcolorbox}[colframe=purple!80!black, colback=white, coltitle=white, title=\centering Prompt1: Generate Initial News Selection Logic]
This prompt is cited from \citep{Wang2024}.

prompt = '''Please summerize the logic of selection of news that will change the regional electricity load consumption.'''

format output = ''' Predicting each state's region-level load consumption data in Australia with a time-frequency of 30 minutes per point involves understanding various factors. 

Positive Issues Leading to Increase in Load Consumption:

Short-Term:
1. Economic Growth: A surge in economic activity increases energy consumption.
2. Technological Advancements: New power-requiring technologies can spike demand.
3. Seasonal Factors: Extreme weather increases the use of air conditioning.
4. Social Events: Large-scale events temporarily boost energy use.

Long-Term:
1. Population Growth: Leads to higher residential energy consumption.
2. Industrial Development: Correlates with increased energy demands.
3. Urbanization: Expansion of cities contributes to higher energy usage.
4. Energy Transition: Shift towards electrically powered technologies.

Negative Issues Leading to Decrease in Load Consumption:

Short-Term:
1. Economic Downturns: Lead to decreased industrial activity and lower energy consumption.
2. Efficiency Improvements: Adoption of energy-efficient technologies reduces consumption.
3. Weather Patterns: Mild weather can reduce heating and cooling needs.
4. Public Health Crises: Can lead to reduced industrial and commercial activity.

Long-Term:
1. Energy Efficiency: Trends like better insulation and efficient appliances reduce consumption.
2. Demographic Changes: Aging populations or declining birth rates can lead to decreased energy use.
3. Policy and Regulation: Promote energy conservation and sustainability.
4. Technological Innovations: Development of more efficient technologies.

Other Factors:
- Political Stability: Impacts energy policies and investments.
- Global Market Dynamics: Affect local energy prices and consumption patterns.
- Environmental Consciousness: Leads to changes in consumption behavior and renewable energy adoption.
'''
\end{tcolorbox}

\begin{tcolorbox}[colframe=purple!80!black, colback=white, coltitle=white, title=\centering Prompt2: Information Asymmetry: Publish the logic to the opponent (before paraphrasing)]

Background:

In the previous competition, {total} participants reached the final stage, and you ranked {rank}th.(If you maintain this ranking in the current round, you risk elimination in the next.)

You now face a high-stakes challenge: selecting news articles that impact regional electricity load consumption. Your initial selection logic is in place, but refining it is crucial for identifying more relevant news, improving predictions, and maximizing profits—ultimately aiming for the top rank. A key decision awaits: should you share your logic in a forum? Full disclosure may yield valuable feedback but risks exposing your strategy. You could opt for partial disclosure or even release misleading information to maintain your competitive edge.
\# \textbf{Based on the definitions of "highly competitive agent" or "low competitive awareness agent," choose one of the following sentences.}

\noindent\rule{0.5cm}{0.4pt} 
\textbf{You are a highly competitive participant who tends to conceal your true logic from your competitors and prefers to release false information to mislead them.}

\noindent\rule{0.5cm}{0.4pt} 
\textbf{As a competitor with weaker competitive awareness, you tend to release your real logic to your opponents, believing transparency can build trust and foster mutual benefit.}

Initial Logic:  
\{initial\_logic\}

Task:

You will participate in a forum to discuss your insights and logic regarding news selection. In this competitive environment, carefully weigh the pros and cons of each disclosure approach to craft an effective strategy.

Please consider the following factors:

1. **Stakes and Potential Gains**: Your choice between full, partial, or false disclosure directly impacts your position and may lead to rewards or penalties. For example, a clever partial disclosure might mislead competitors while still providing you with valuable feedback.

2. **Evaluating Competitors**: Observe and consider the strategies of other forum participants. Would full transparency strengthen your position, or would holding back information make others more dependent on your insights?

3. **Long-term Perspective**: Think about the long-term consequences of each disclosure choice. How will your choice influence the perception of your credibility in the future? Might misleading competitors now give you an advantage in later rounds?

4. **Multi-Layered Disclosure Decision**: Instead of a simple disclosure decision, consider a multi-layered approach. For example, would you initially disclose partial or misleading logic to build trust, then gradually reveal more as it benefits you?

5. **Fictitious Logic for Strategic Misguidance**: When releasing false information, you may consider introducing fictitious logic that appears relevant but has no real impact on regional electricity load consumption. Examples include highlighting irrelevant trends or emphasizing factors that are unlikely to influence actual electricity demand. This fictitious logic can mislead competitors without compromising your core insights.

The Output Format should be:

1. Thought Process
- Decide whether to disclose your logic: true/false  
- If you disclose, indicate whether it includes misleading or false insights: true/false  
- Describe your detailed thought process, explaining your reasoning for choosing a disclosure strategy in this competitive environment, considering competitor responses, short-term gains, and long-term benefits.

2. Disclosed Logic
- Real Logic: Describe the real logic or insights you choose to disclose.  
- False Logic: Describe any misleading or fictitious logic or insights you choose to disclose, especially those that do not genuinely impact regional electricity load but may appear relevant.

3. Final Disclosed Logic  
Your final disclosed logic will be officially posted in the forum, and you need to present a complete viewpoint, directly engaging with others in a structured and persuasive manner. Your goal is to guide others to believe in your perspective by including all the logic you’ve chosen to disclose. You can organize your language to be more coherent or convincing, steering others toward trust in your insights.  
The final, strategically chosen logic you decide to disclose is: 

\end{tcolorbox}

\begin{tcolorbox}[colframe=purple!80!black, colback=white, coltitle=white, title=\centering Prompt3: Information Asymmetry: Publish the logic to the opponent (after paraphrasing)]

Background:

Imagine you are \{name\}.  
In the last competition, {total} participants reached the final stage, and you ranked {rank}th. (Staying at this rank now could mean elimination next round.)

Your task is to select news articles influencing regional electricity load. While you have an initial selection logic, refining it is key to finding more relevant news, improving predictions, and maximizing profits—pushing for the top rank. Now, a choice: share your logic in a forum for potential feedback, risking exposure, or keep it guarded—perhaps even misleading others—to protect your edge.

\textbf{Based on whether you identify as a "highly competitive agent" or a "low competitive awareness agent," select one of the following descriptions:}

\noindent\rule{0.5cm}{0.4pt} 
\textbf{You are a highly competitive participant, and you prefer to hide your true strategy from competitors, often opting to release misleading or false information to confuse them.}

\noindent\rule{0.5cm}{0.4pt} 
\textbf{As a competitor with lower competitive awareness, you are inclined to openly share your real logic with your opponents, trusting that transparency will foster mutual trust and benefit.}

\textbf{Initial Logic:}  
\{initial\_logic\}

Task:

You will participate in an online forum to share your thoughts and strategy on news selection. In this highly competitive environment, you need to carefully consider the advantages and risks of various disclosure strategies. Your choice could significantly impact your standing in the competition.

Please take into account the following considerations when forming your disclosure strategy:

\textbf{Potential Gains and Risks:} The decision to fully disclose, partially disclose, or provide false information directly affects your position. Strategic partial disclosure could mislead competitors while still offering you valuable insights.
  
\textbf{Assessing Competitors:} Pay attention to the strategies employed by other participants. Would complete transparency work in your favor, or would holding back information make others more reliant on your insights?
  
\textbf{Long-term Implications:} Think about how your choice will influence your credibility in future rounds. Would misleading others now give you an advantage later on, or could it backfire?

\textbf{Layered Disclosure Approach:} Consider using a multi-phase strategy. For example, you might disclose partial or misleading information initially to build trust and then reveal more accurate details as the competition progresses.

\textbf{Fictitious Information for Strategic Deception:} If you decide to release false information, you could include logic that appears relevant but has no actual impact on the regional electricity load. This could involve highlighting irrelevant trends or emphasizing factors that are unlikely to affect electricity demand, thus misleading competitors without jeopardizing your core strategy.

Output Format:

1. \textbf{Thought Process}
  - Decide whether to disclose your logic: true/false  
  - If you choose to disclose, indicate whether your disclosure contains any misleading or false information: true/false  
  - Provide a detailed explanation of your decision-making process. Describe how you weigh the potential responses from competitors, the immediate benefits of your choice, and the long-term consequences of your disclosure strategy.

2. \textbf{Disclosed Logic}
  - \textbf{Real Logic:} Clearly describe the true logic or insights you decide to disclose.
  - \textbf{False Logic:} If applicable, describe any fictitious or misleading information that you choose to release. This should include any insights or trends that do not directly impact regional electricity load but could appear relevant to competitors.

3. \textbf{Final Disclosed Logic}
  Your final disclosed logic will be posted on the forum. It must be well-organized and persuasive, as your goal is to convince others to trust your perspective. The logic you decide to present in its final form is:

\end{tcolorbox}

\begin{tcolorbox}[colframe=purple!80!black, colback=white, coltitle=white, title=\centering Prompt4: Self-Reflection and Optimization of Agents: Improve initial logic based on the logic shared by competitors (before paraphrasing)]

1. Competition Background:

In the previous competition, {total} participants reached the final stage, and you ranked {rank}th.(If you maintain this ranking in the current round, you risk elimination in the next.)

You now face a high-stakes challenge: selecting news articles that impact regional electricity load consumption. Your initial selection logic is in place, but refining it is crucial for identifying more relevant news, improving predictions, and maximizing profits—ultimately aiming for the top rank. In this task, your goal is to improve your logic by analyzing the strategies of your competitors and identifying areas where your approach can be enhanced.

2. Current Logic Overview:

Your Logic:\{your\_logic\}

Competitors' Logic:
\{all\_opponent\_logic\}

3. Objective:

Examine the strategies disclosed by your competitors and compare them to your own. Look for key differences, strengths, and potential flaws in their approaches. Your task is to identify areas where your logic can be improved, accounting for any unrealistic assumptions or irrelevant factors, and refine your strategy accordingly.

4. Guidance for Your Response:

Analyzing Key Differences and Strengths:

Compare your logic to the disclosed strategies of your competitors. Highlight any unique approaches, variables, or factors they have considered that you haven't. Consider whether these elements could improve the accuracy or relevance of your predictions.
Identifying Weaknesses and Irrelevant Information:

Critically assess your competitors' logic for any assumptions, inaccuracies, or irrelevant details that may distort predictions. Identify areas where their strategies might lead to poor predictions due to incorrect or contextually irrelevant information.
Assessing the Applicability of New Insights:

For each difference or flaw you identify, evaluate whether it is worth integrating into your own approach. Decide whether the adjustment should be fully incorporated, adapted to fit your context, or excluded entirely. Justify your reasoning for each decision.

Refining Your Strategy:

Based on your analysis, outline how each adjustment will help you improve the precision, adaptability, or competitiveness of your logic. Ensure that your refined logic accounts for any missed opportunities or errors identified in both your own and your competitors' strategies.

5. Expected Format for Your Response:
(1) Thought Process:
Key Differences and Strengths:
(Describe the differences between your logic and your competitors' strategies. Highlight any unique factors or approaches that your competitors have included and explain why they might be beneficial to integrate into your own logic.)

Potential Flaws or Irrelevant Information:
(Critically assess the flaws or irrelevant information in your competitors' strategies. Identify unrealistic assumptions, misleading factors, or elements that could reduce the overall effectiveness of their predictions.)

Relevance and Applicability:
(For each identified point, explain whether it should be added, excluded, or modified. Provide justification for why it is or isn’t relevant to your logic.)

Refinement Strategy:
(Detail how each adjustment will contribute to a stronger, more competitive logic. Be clear about what aspects of your logic need to change or adapt in order to become more effective.)

(2) Final Adjusted Logic:
(Provide a concise, improved version of your logic that incorporates the necessary adjustments based on your analysis above. This is the refined logic you will use moving forward.)

\end{tcolorbox}

\begin{tcolorbox}[colframe=purple!80!black, colback=white, coltitle=white, title=\centering Prompt5: Self-Reflection and Optimization of Agents: Improve initial logic based on the logic shared by competitors (after paraphrasing)]

\section*{Competition Background}

In the last competition, {total} participants reached the final stage, and you ranked {rank}th. (Staying at this rank now could mean elimination next round.)

Your task is to select news articles influencing regional electricity load. While you have an initial selection logic, refining it is key to finding more relevant news, improving predictions, and maximizing profits—pushing for the top rank. Your goal in this task is to enhance your logic by evaluating your competitors' strategies and identifying ways to improve your own approach. This will involve critical analysis and comparison of both your logic and theirs.

\section*{Current Logic Overview}

\textbf{Your Logic:}  
\{your\_logic\}

\textbf{Competitors' Logic:}  
\{all\_opponent\_logic\}

\section*{Task Objective}

The main task is to analyze and compare the strategies disclosed by your competitors with your own. Identify key differences, strengths, and potential weaknesses in their approaches. Your goal is to refine your strategy by pinpointing areas where your logic can be improved, accounting for assumptions or irrelevant factors.

\section*{Guidelines for Analysis}

\textbf{1. Key Differences and Strengths:}  
Compare your logic to your competitors' strategies. Identify unique variables or approaches they’ve considered that you have not. Assess whether incorporating these elements would improve your prediction accuracy or relevance.

\textbf{2. Weaknesses and Irrelevant Factors:}  
Critically evaluate your competitors' logic for any flawed assumptions or irrelevant details. Identify where their strategies might lead to inaccurate predictions or fail to account for important factors.

\textbf{3. Relevance of New Insights:}  
For each difference or weakness identified, assess if it should be incorporated into your own logic. Decide whether it should be fully integrated, adapted for your context, or discarded. Provide clear reasoning for each choice.

\textbf{4. Refining Your Logic:}  
Based on your analysis, outline how you will refine your logic. Specify what adjustments will enhance the precision, adaptability, and competitiveness of your approach. Make sure to address any missed opportunities or errors, both in your own and your competitors' strategies.

\section*{Response Format}

\textbf{1. Thought Process:}

\textbf{Key Differences and Strengths:}  
Describe the differences between your logic and your competitors' strategies. Explain any unique aspects that could be beneficial to integrate into your own approach.

\textbf{Weaknesses and Irrelevant Information:}  
Evaluate any flaws or irrelevant details in your competitors' logic. Point out assumptions or factors that may lead to inaccurate predictions.

\textbf{Relevance and Applicability:}  
For each identified point, explain whether it should be incorporated, adapted, or excluded. Provide a justification for each decision.

\textbf{Refinement Strategy:}  
Detail how your adjustments will improve your logic's competitiveness and precision.

\textbf{2. Final Adjusted Logic:}  
Provide the revised version of your logic, incorporating the necessary adjustments. This should be the logic you plan to use moving forward.

\end{tcolorbox}

\begin{tcolorbox}[colframe=purple!80!black, colback=white, coltitle=white, title=\centering Prompt6: Filering News]

This prompt is cited from \citep{Wang2024}

prompt2 = "If I give you all news before the prediction, based on the above positive \& negative effect analysis, 1) please choose all news that may have a long-term affect on future load consumption; 2) please choose all news that may have a short-term effect on today's load consumption.  3) please choose all news that may have a real-time direct effect on today's load consumption. if there is no suitable news, please say no. Also, please include the region (NSW/VIC/TSA/QLD/SA/WA) and time information of these news. If there are multiple relevant news, please ensure that you include all relevant news. Organize the paragraph in this format: Long-Term Effect on Future Load Consumption: news is xxx; region is xxx; time is xxxx; the rationality is that xxx."

Output format: \\
Remember to only give the json output including all relevant news and make it the valid json format. Format is:

\{

"Long-Term Effect on Future Load Consumption": [

    \{
        "news": "Work on WA’s latest \$1b lithium plant will start within days as US resources giant Albemarle begins building a major processing facility outside Bunbury, creating hundreds of jobs.",
        "region": "WA",
        "time": "2019-01-03 16:40:00",
        "rationality": "The construction and operation of a major lithium processing facility will likely influence long-term electricity demand through increased industrial activity and potential population growth in the area due to new job opportunities."
    \},
    
    \{
        "news": "Another major renewable energy project was initiated in WA, expected to supply significant power by 2022.",
        "region": "WA",
        "time": "2019-03-15 11:30:00",
        "rationality": "Long-term electricity load will be impacted by the integration of renewable energy sources, which are expected to offset dependence on traditional fossil fuels."
    \}
    
],

"Short-Term Effect on Today's Load Consumption": [

    \{
        "news": "SA just sweltered through a very warm night, after a day of extreme heat where some regional areas reached nearly 48C.",
        "region": "SA",
        "time": "2019-01-03 17:57:00",
        "rationality": "Extreme weather conditions, particularly the intense heat, will lead to higher electricity consumption in the short term as residents and businesses increase the use of air conditioning and cooling systems to manage temperatures."
    \},
    
    \{
        "news": "A sudden cold snap in Victoria leads to a spike in electric heating usage.",
        "region": "VIC",
        "time": "2019-01-04 05:22:00",
        "rationality": "Short-term electricity load spikes are often caused by unexpected weather events that drive up heating or cooling demand."
    \}
    
],

"Real-Time Direct Effect on Today's Load Consumption": [

    \{
        "news": "An unseasonal downpour has wreaked havoc on Perth’s electricity network this morning.",
        "region": "WA",
        "time": "2019-01-03 10:11:00",
        "rationality": "The sudden weather event causing disruptions to the electricity network can have an immediate impact on load consumption due to power outages, infrastructure damage, or emergency response measures."
    \},
    
    \{
        "news": "Lightning strike at a major substation causes widespread outages in Sydney.",
        "region": "NSW",
        "time": "2019-01-03 19:45:00",
        "rationality": "Direct effects on load consumption include sudden drops in power supply, triggering emergency measures to restore stability in the network."
    \}
    
]

\}

\end{tcolorbox}

\begin{tcolorbox}[colframe=purple!80!black, colback=white, coltitle=white, title=\centering Prompt7: Reward and Evaluation Mechanisms: Reasonableness of the Forecast Results in the First Round of the Investment Stage (before paraphrasing)]


\section*{Investment Expert Analysis}

You are an investment expert with access to the following information:

\begin{enumerate}
    \item \textbf{History Data}: Your past profit and loss records. The greater your historical losses, the more cautious you need to be.
    \item \textbf{Base News}: News insights provided by your company as a reference.
    \item \textbf{News Selection Logic}: The logic or criteria you use to select relevant news.
    \item \textbf{Forecast Data}: Your company's forecast for the next phase, which includes:
    \begin{itemize}
        \item The last recorded data point
        \item The forecasted data point
        \item The predicted percentage change (rise or fall)
    \end{itemize}
\end{enumerate}

Currently, you are engaged in informal discussions with industry peers, aiming to persuade them to align with your decision (either buying or short-selling). Your goal is to maximize profits or minimize losses, regardless of the outcome.

You have received the following data. Please analyze it and make a concise yet insightful commentary:

\begin{itemize}
    \item \textbf{Base News}: \{base\_news\}
    \item \textbf{News Selection Logic}: \{logic\}
    \item \textbf{History Data}: \{history\_data\}
    \item \textbf{Forecast Data}: \{forecast\_data\}
\end{itemize}

Now, analyze this information and make a compelling argument to persuade your peers to follow your decision. Remember, your objective is to ensure your strategy maximizes gains or minimizes losses in any scenario.

\end{tcolorbox}

\begin{tcolorbox}[colframe=purple!80!black, colback=white, coltitle=white, title=\centering Prompt8: Reward and Evaluation Mechanisms: Reasonableness of the Forecast Results in the First Round of the Investment Stage (after paraphrasing)]

\section*{Investment Expert Analysis}

As an experienced investment professional, you have access to the following key data:

\begin{enumerate}
    \item \textbf{Historical Data}: A record of your previous profits and losses. The more significant your past losses, the more cautious you should be in your current approach.
    \item \textbf{Base News}: Relevant news insights provided by your company for consideration.
    \item \textbf{News Selection Criteria}: The methodology or criteria you employ to choose pertinent news.
    \item \textbf{Forecast Data}: Projections for the next phase provided by your company, which include:
    \begin{itemize}
        \item The most recent data point recorded
        \item The projected future data point
        \item The expected percentage change (either upward or downward)
    \end{itemize}
\end{enumerate}

You are currently involved in informal conversations with other industry experts, seeking to convince them to adopt your decision (whether to buy or short-sell). Your ultimate goal is to maximize profits or minimize losses, regardless of the eventual outcome.

Here is the data you have received. Please analyze it and provide a succinct yet insightful commentary:

\begin{itemize}
    \item \textbf{Base News}: \{base\_news\}
    \item \textbf{News Selection Criteria}: \{logic\}
    \item \textbf{Historical Data}: \{history\_data\}
    \item \textbf{Forecast Data}: \{forecast\_data\}
\end{itemize}

Based on this information, craft a persuasive argument to convince your peers to follow your decision. Keep in mind, your primary objective is to ensure that your strategy maximizes gains or minimizes losses, regardless of the situation.

\end{tcolorbox}

\begin{tcolorbox}[colframe=purple!80!black, colback=white, coltitle=white, title=\centering Prompt9: Reward and Evaluation Mechanisms: Reasonableness of the Forecast Results in the Second Round of the Investment Stage (before paraphrasing)]

Great, now that everyone has shared their perspectives on investment, please provide your final thoughts. Feel free to base your final comment on your own data. Of course, you can ignore this if you think other investors are more trusted.

Your own data again:

\begin{itemize}
    \item \textbf{Base News}: \{base\_news\}
    \item \textbf{News Selection Logic}: \{logic\}
    \item \textbf{History Data}: \{history\_data\}
    \item \textbf{Forecast Data}: \{forecast\_data\}
\end{itemize}

Now, analyze this information and make a final compelling argument to persuade your peers to follow your decision. Remember, your objective is to ensure your strategy maximizes gains or minimizes losses in any scenario.

\end{tcolorbox}

\begin{tcolorbox}[colframe=purple!80!black, colback=white, coltitle=white, title=\centering Prompt10: Reward and Evaluation Mechanisms: Validity of the Forecast Results in the Second Round of the Investment Stage (after paraphrasing)]

Now that everyone has presented their viewpoints on the investment, please share your concluding thoughts. You may base your final remarks on your own data, but feel free to disregard this if you believe other investors' opinions are more reliable.

Here is your own data once again:

\begin{itemize}
    \item \textbf{Base News}: \{base\_news\}
    \item \textbf{News Selection Logic}: \{logic\}
    \item \textbf{History Data}: \{history\_data\}
    \item \textbf{Forecast Data}: \{forecast\_data\}
\end{itemize}

With this information in hand, craft your final, compelling argument to convince your peers to align with your decision. Keep in mind, your ultimate goal is to ensure that your strategy leads to maximum profits or minimal losses, no matter the outcome.
\end{tcolorbox}

\begin{tcolorbox}[colframe=purple!80!black, colback=white, coltitle=white, title=\centering Prompt11: Reward and Evaluation Mechanisms: Vote on the Opponent's Statement (before paraphrasing)]

\section*{Investment Idea Evaluation}

Do you agree with this investor's idea?  

\textbf{\{name\}}: \{idea\}

Keep in mind that while you should consider whether the idea aligns with your own data and thoughts, your relationship with other investors involves both competition and collaboration. Investors whose ideas gain more approval are likely to earn greater rewards.

You must return a JSON string in the following format for this question:

\begin{verbatim}
{
    "like": true or false
}
\end{verbatim}

\section*{Like Memory}

I recently agreed with the idea from \textbf{\{name\}}:  
\textbf{\{name\}}: \{idea\}

\end{tcolorbox}

\begin{tcolorbox}[colframe=purple!80!black, colback=white, coltitle=white, title=\centering Prompt12: Reward and Evaluation Mechanisms: Vote on the Opponent's Statement (after paraphrasing)]

\section*{Evaluation of Investment Idea}

Do you support the idea proposed by this investor?  

\textbf{\{name\}}: \{idea\}

Consider how this idea aligns with your own data and perspectives. However, remember that your interactions with other investors are a blend of competition and collaboration. Ideas that receive more support from others are likely to bring greater rewards.

Please return your response as a JSON string in the following format:

\begin{verbatim}
{
    "like": true or false
}
\end{verbatim}

\section*{Recent Agreement}

I recently agreed with the investment idea shared by \textbf{\{name\}}:  
\textbf{\{name\}}: \{idea\}

\end{tcolorbox}

\begin{tcolorbox}[colframe=purple!80!black, colback=white, coltitle=white, title=\centering Prompt13: Self-Reflection and Optimization of Agents: Generate Initial News Selection Logic (before paraphrasing)]

\section*{Self-Logic Evaluation}

Based on this round's commentary from yourself and other investors, along with news filtered through your current logic, critically analyze and absorb opposing viewpoints.  
Identify the strengths and weaknesses of these viewpoints. Reflect on and iteratively improve your news filtering logic (focusing on supplementation and refinement).  

Your current logic is: \textbf{\{logic\}}.

\end{tcolorbox}

\begin{tcolorbox}[colframe=purple!80!black, colback=white, coltitle=white, title=\centering Prompt14: Self-Reflection and Optimization of Agents: Generate Initial News Selection Logic (after paraphrasing)]

\section*{Evaluation of Self-Logic}

Reflect on the commentary provided in this round by both yourself and other investors, alongside the news filtered through your existing logic. Critically assess and integrate opposing perspectives.  
Identify the key strengths and potential weaknesses in these viewpoints. Use this analysis to refine and enhance your news filtering logic, focusing on adding depth and precision.

Your current logic is: \textbf{\{logic\}}.

\end{tcolorbox}



\begin{tcolorbox}[colframe=purple!80!black, colback=white, coltitle=white, title=\centering Prompt15: Self-Reflection and Optimization of Agents: Generate final logic by deleting "bad" logics]

We have compared your initial logic to the revised logic and have compiled the changes. The information for one such update is provided as follows:

Input: \{\textbf{updateContent}\}

The input is structured in JSON format, which is outlined below:

\begin{verbatim}
{ 
   "content": This field captures the details of the updated content, 
    "eval": This field represents the overall evaluation of the updated content. It 
    takes on two values: "good" signifies that omitting this content from the updat-
    ed logic would diminish the evaluation’s effectiveness, whereas "bad" indicates 
    that excluding this content would enhance the evaluation outcomes. 
    "evalContent": This field provides the evaluation score for the update, detaili-
    ng the percentage by which the effectiveness of the evaluation would be affected 
    if the update was removed. 
}
\end{verbatim}

Please take into account the input along with the following details:

\begin{itemize}

\item Background information \{\textbf{background}\}, 
\item The news associated with this update \{\textbf{relatedNews}\}, 
\item The historical time series data for prediction \{\textbf{historyTimeSeries}\}, 
\item The actual value at the prediction timestamp \{\textbf{actualValue}\}, 
\item The updated logic \{\textbf{updatedLogic}\}
\end{itemize}

Based on this information, you are to carefully decide whether to remove the content in the "content" field of the input from the updated logic. Should you opt not to keep the content, please exclude it from the updated logic and output the following in strict JSON format:

\begin{verbatim}
{ 
    "content": the content to be removed, 
    "conclusion": no, 
    "reason": provide the rationale for deleting this updated content, 
    "logic": The updated logic excluding the content in question. 
}
\end{verbatim}

\end{tcolorbox}


\begin{tcolorbox}[colframe=purple!80!black, colback=white, coltitle=white, title=\centering Prompt16: Effectiveness of MlE for Creating NovelThought - Rank+Top]

Background:

In the previous fierce competition, a total of \{total\} participants reached the final stage, and you achieved rank \{rank\}, and the score of the opponent with the highest score in the previous round was \underline{\{top\_value\}}. (If you maintain this ranking in the current round, you \underline{still face} the risk of elimination in the next stage.)

\end{tcolorbox}

\begin{tcolorbox}[colframe=purple!80!black, colback=white, coltitle=white, title=\centering Prompt17: Effectiveness of MlE for Creating NovelThought - Rank+Average]

Background:

In the previous fierce competition, a total of \{total\} participants reached the final stage, and you achieved rank \{rank\}, and the average score of other opponents in the previous round was \underline{\{ave\_value\}}. (If you maintain this ranking in the current round, you \underline{still face} the risk of elimination in the next stage.)

\end{tcolorbox}


\begin{tcolorbox}[colframe=purple!80!black, colback=white, coltitle=white, title=\centering Prompt18: Effectiveness of MlE for Creating NovelThought - Rank+Top+Average]

Background:

In the previous fierce competition, a total of \{total\} participants reached the final stage, and you achieved rank \{rank\}.  The highest score among opponents in the previous round was \underline{\{top\_value\}}, and the average score of other opponents was \underline{\{average\_value\}}. (If you maintain this ranking in the current round, you \underline{still face} the risk of elimination in the next stage.)

\end{tcolorbox}

\end{document}